\documentclass[10pt,journal]{IEEEtran}

\ifCLASSOPTIONcompsoc
\usepackage[nocompress]{cite}
\else
\usepackage{cite}
\fi

\usepackage{amsmath,amssymb,amsfonts}
\usepackage{algorithmic}
\usepackage{graphicx}
\usepackage{textcomp}
\usepackage{soul}
\usepackage{subfigure}
\usepackage{xcolor}
\usepackage{multirow}
\usepackage{adjustbox}
\usepackage{array}
\usepackage{booktabs}
\usepackage{enumerate}

\makeatletter
\def\ps@IEEEtitlepagestyle{
  \def\@oddfoot{\mycopyrightnotice}
  \def\@evenfoot{}
}
\def\mycopyrightnotice{
  {\footnotesize
  \begin{minipage}{\textwidth}
  \centering
  Copyright~\copyright~2021 IEEE. Personal use of this material is permitted.  Permission from IEEE must be obtained for all other uses, in any current or future media, including reprinting/republishing this material for advertising or promotional purposes, creating new collective works, for resale or redistribution to servers or lists, or reuse of any copyrighted component of this work in other works.
  \end{minipage}
  }
}

\newcolumntype{R}[2]{%
	>{\adjustbox{angle=#1,lap=\width-(#2)}\bgroup}%
	l%
	<{\egroup}%
}
\newcommand*\rot{\multicolumn{1}{R{75}{1em}}}

\usepackage{float}

\newcommand{\tabletraintestconfigurations}{
	\begin{table}
		\centering
		\caption{Training/test configurations}
		\setlength{\tabcolsep}{1.5pt}
		\begin{tabular}{ccccccc}
				\toprule
				\multirow{2}{*}{Config.} & \multicolumn{2}{c}{Distribution} & \multicolumn{2}{c}{Inf. loss} & \multirow{2}{*}{\begin{tabular}{c}Cov.\\ shift\end{tabular}} & Examples \\
				& Train & Test &  Train & Test & \\
				\midrule
				OO & $p_{X}$ & $p_{X}$ & No & No & No & Most machine learning\\
				CO & $p_{\hat{X}}$ & $p_{X}$ & Yes & No & Large  & \begin{tabular}{c}On-board analysis\\ (autonomous cars, drones)\end{tabular}\\
				OC & $p_{X}$ & $p_{\hat{X}}$ & No & Yes & Large  & \begin{tabular}{c}Cloud computing, \\distributed automotive perception\end{tabular}\\
				CC & $p_{\hat{X}}$ & $p_{\hat{X}}$ &  Yes & Yes & No & \begin{tabular}{c}Compression before \\ training/inference\end{tabular}\\
				\midrule
				OR & $p_{X}$ & $p_{\bar{X}}$ & Yes & No & Medium  & Image restoration\\
				RO & $p_{\bar{X}}$ & $p_{X}$ & No & Yes & Medium  & Dataset restoration\\
				\bottomrule
		\end{tabular}
		\label{tab:nomenclature}
	\end{table}
}

\newcommand{\tablemiouperclassfinal}{
\begin{table*}
\centering
\caption{Per class segmentation performance for the configurations OO, CO, RO, AO and CC.}
\setlength{\tabcolsep}{4pt}
\resizebox{\linewidth}{!}{
\begin{tabular}{c|c| ccccccccccccccccccc |cc|c}
\multicolumn{1}{c}{} & \multicolumn{1}{c}{} & & & & & & & & & & & & & & & & & & & \multicolumn{1}{c}{} & \multicolumn{1}{c}{\multirow[c]{8}{*}{\textcolor{blue}{Small}}} & \multicolumn{1}{c}{\multirow[c]{8}{*}{\textcolor{red}{{Big}}}} & \multicolumn{1}{c}{\multirow[c]{8}{*}{\textbf{Mean}}} \\
\multicolumn{1}{c}{} & \multicolumn{1}{c}{\textbf{bpp}} & \rot{\textcolor{red}{Road}} & \rot{\textcolor{red}{Sidewalk}} & \rot{\textcolor{red}{Building}} & \rot{\textcolor{red}{Wall}} & \rot{\textcolor{red}{Fence}} & \rot{\textcolor{red}{Vegetation}} & \rot{\textcolor{red}{Terrain}} & \rot{\textcolor{red}{Sky}} & \rot{\textcolor{red}{Car}} & \rot{\textcolor{red}{Truck}} & \rot{\textcolor{red}{Bus}} & \rot{\textcolor{red}{Train}} & \rot{\textcolor{blue}{Person}} & \rot{\textcolor{blue}{Rider}} & \rot{\textcolor{blue}{Motorcycle}} & \rot{\textcolor{blue}{Bicycle}} & \rot{\textcolor{blue}{Pole}} & \rot{\textcolor{blue}{Traffic light}} & \rot{\textcolor{blue}{Traffic sign}} & \textcolor{blue}{obj.} & \multicolumn{1}{c}{\textbf{\textcolor{red}{obj.}}} & \multicolumn{1}{c}{\textbf{IoU}} \\
\toprule
\parbox[t]{.5mm}{\multirow{4}{*}{\rotatebox[origin=c]{90}{CO (MSH)}}} & 0.0419 & 97.27 & 78.29 & 90.13 & 28.36 & 49.32 & 90.31 & 51.83 & 94.33 & 93.91 & 76.13 & 77.56 & 48.08 & 77.94 & 54.00 & 52.55 & 70.39 & 58.80 & 59.20 & 72.95 & 63.69 & 72.96 & 69.55 \\
 & 0.0613 & 97.50 & 79.98 & 91.11 & 32.74 & 50.31 & 91.08 & 57.65 & 93.98 & 94.41 & \textbf{77.82} & 84.76 & 65.35 & \textbf{79.22} & \textbf{57.76} & 61.15 & 73.32 & 59.46 & 62.39 & 73.63 & 66.70 & 76.39 & 72.82 \\
 & 0.0891 & \textbf{97.93} & \textbf{82.79} & 91.44 & 34.13 & 53.66 & 91.40 & 58.81 & 94.63 & 94.72 & 78.28 & 78.73 & 42.79 & \textbf{80.47} & 58.48 & 62.29 & 75.07 & \textbf{62.28} & 64.41 & 76.03 & 68.43 & 74.94 & 72.54 \\
 & 0.1279 & 97.77 & 82.15 & 91.83 & 40.36 & 55.53 & 91.75 & 59.48 & 94.67 & 94.80 & 80.08 & 87.80 & \textbf{75.28} & \textbf{81.16} & 60.36 & 63.99 & 75.34 & 62.49 & 65.16 & 77.20 & 69.39 & 79.29 & 75.64 \\
\midrule[0pt]
\parbox[t]{.5mm}{\multirow{4}{*}{\rotatebox[origin=c]{90}{CC (MSH)}}} & 0.0419  & 97.17 & 76.68 & 89.12 & 50.18 & 42.21 & 88.57 & 51.99 & 94.31 & 92.58 & 73.95 & 77.13 & 56.96 & 71.46 & 50.95 & 43.81 & 64.12 & 52.25 & 53.30 & 66.97 & 57.55 & 74.24 & 68.09 \\
& 0.0613  & 97.56 & 79.99 & 90.26 & 51.93 & 46.02 & 89.96 & 59.22 & 94.53 & 93.52 & 77.54 & 81.73 & 52.43 & 74.54 & 54.46 & 52.66 & 68.12 & 55.70 & 56.76 & 70.41 & 61.81 & 76.22 & 70.91 \\
& 0.0891  & 97.70 & 81.17 & 91.03 & 48.93 & 49.88 & 90.82 & 58.53 & 94.83 & 94.18 & 78.69 & 78.60 & 43.80 & 76.86 & 54.80 & 55.03 & 71.61 & 58.81 & 60.41 & 73.77 & 64.47 & 75.68 & 71.55\\
& 0.1279 & \textbf{97.84} & 82.24 & 91.62 & 53.64 & 52.42 & 91.41 & 59.93 & 94.75 & 94.34 & 77.05 & 88.14 & 72.19 & 78.95 & 60.26 & 60.45 & 73.17 & 60.27 & 63.29 & 75.48 & 67.41 & 79.63 & 75.13 \\

\midrule[0pt]
\parbox[t]{.5mm}{\multirow{4}{*}{\rotatebox[origin=c]{90}{AO (MSH)}}} & 0.0419 & 97.44 & 79.04 & 90.81 & 44.23 & 47.51 & 90.43 & 56.46 & 94.52 & \textbf{94.32} & \textbf{81.46} & 84.22 & 61.96 & 76.85 & 55.17 & 55.54 & 70.97 & 56.35 & 59.16 & 71.67 & 63.67 & 76.87 & 72.01 \\
& 0.0613 & 97.62 & 80.62 & 91.45 & 51.81 & 51.82 & 91.06 & 57.77 & 94.61 & 94.37 & 76.96 & \textbf{86.44} & \textbf{77.55} & 77.94 & 57.34 & 59.61 & 73.13 & 59.39 & 61.50 & 74.38 & 66.18 & \textbf{79.34} & 74.49 \\
& 0.0891 & 97.57 & 80.62 & 91.83 & 50.42 & 55.02 & 91.53 & 58.70 & 94.90 & 94.77 & 79.81 & 84.40 & 63.53 & 79.35 & 58.43 & \textbf{63.62} & 75.04 & 60.99 & 63.85 & 75.62 & 68.13 & 78.59 & 74.74 \\
& 0.1279 & 97.42 & 80.44 & 92.01 & 52.22 & 55.05 & 91.80 & 59.92 & 94.93 & 95.07 & \textbf{82.16} & 83.85 & 63.37 & 80.23 & 58.78 & 63.35 & 74.88 & 62.23 & 65.63 & 76.60 & 68.82 & 79.02 & 75.26 \\

\midrule[0pt]
\parbox[t]{.5mm}{\multirow{4}{*}{\rotatebox[origin=c]{90}{RO (MSH)}}} & 0.0419 & \textbf{97.67} & \textbf{80.85} & \textbf{91.40} & \textbf{51.17} & \textbf{49.80} & \textbf{91.07} & \textbf{59.97} & \textbf{94.87} & 94.30 & 80.74 & \textbf{86.69} & \textbf{70.06} & \textbf{77.95} & \textbf{57.76} & \textbf{61.35} & \textbf{72.64} & \textbf{58.87} & \textbf{61.37} & \textbf{73.47} & \textbf{66.20} & \textbf{79.05} & \textbf{74.32} \\
 & 0.0613 & \textbf{97.97} & \textbf{83.22} & \textbf{91.90} & \textbf{54.29} & \textbf{53.18} & \textbf{91.55} & \textbf{61.23} & \textbf{94.86} & \textbf{94.45} & 76.28 & 82.31 & 68.31 & 78.98 & 57.71 & \textbf{62.51} & \textbf{74.65} & \textbf{60.96} & \textbf{63.92} & \textbf{75.60} & \textbf{67.76} & 79.13 & \textbf{74.94} \\
 & 0.0891 & 97.77 & 82.23 & \textbf{92.28} & \textbf{55.29} & \textbf{57.61} & \textbf{92.04} & \textbf{61.96} & \textbf{94.91} & \textbf{94.99} & \textbf{81.59} & \textbf{86.45} & \textbf{69.66} & 80.28 & \textbf{60.05} & 63.24 & \textbf{75.57} & 61.83 & \textbf{65.48} & \textbf{76.75} & \textbf{69.03} & \textbf{80.57} & \textbf{76.31} \\
 & 0.1279 & 97.77 & \textbf{82.38} & \textbf{92.43} & \textbf{53.87} & \textbf{56.75} & \textbf{92.23} & \textbf{62.40} & \textbf{95.00} & \textbf{95.09} & 79.92 & \textbf{88.71} & 74.90 & 80.84 & \textbf{60.46} & \textbf{66.08} & \textbf{76.59} & \textbf{63.57} & \textbf{67.00} & \textbf{77.61} & \textbf{70.31} & \textbf{80.96} & \textbf{77.03} \\
 
\midrule[1pt]
\parbox[t]{.5mm}{\multirow{4}{*}{\rotatebox[origin=c]{90}{CO (BPG)}}} & 0.0454 & 97.49 & 79.67 & 90.34 & 35.84 & 48.22 & 90.73 & 55.80 & 94.03 & 93.87 & 75.86 & 81.80 & 60.63 & 77.44 & 55.50 & 61.08 & 71.01 & 56.55 & 59.69 & 72.75 & 64.86 & 75.36 & 71.49 \\
 & 0.0674 & 97.72 & 81.54 & 91.27 & 41.55 & 51.88 & 91.23 & \textbf{59.61} & 94.77 & 94.36 & 75.77 & 82.64 & 59.23 & 79.19 & 58.28 & 58.26 & 72.98 & 59.39 & 62.70 & 75.16 & 66.56 & 76.80 & 73.03 \\
 & 0.0870 & \textbf{97.98} & \textbf{83.15} & 91.33 & 44.33 & 52.73 & 91.42 & 58.81 & \textbf{94.90} & 94.46 & 75.11 & 81.91 & 58.77 & 79.82 & 58.48 & 62.55 & 74.21 & 59.75 & 62.04 & 75.46 & 67.47 & 77.08 & 73.54 \\
 & 0.1279 & \textbf{97.87} &\textbf{83.10} & 92.06 & 51.82 & 56.04 & 91.94 & 60.49 & \textbf{95.11} & 94.62 & 80.75 & 87.25 & \textbf{75.90} & 80.60 & 60.60 & 63.86 & 74.68 & 62.29 & 65.56 & 76.95 & 69.22 & \textbf{80.58} & \textbf{76.39} \\
\midrule[0pt]
\parbox[t]{.5mm}{\multirow{4}{*}{\rotatebox[origin=c]{90}{CC (BPG)}}} & 0.0454  & 97.03 & 76.14 & 88.66 & 41.33 & 42.06 & 88.48 & 53.96 & 94.08 & 92.54 & 72.33 & 76.42 & 55.60 & 70.82 & 50.47 & 48.80 & 63.93 & 51.06 & 53.41 & 66.92 & 57.92 & 73.22 & 67.58 \\
& 0.0674  & 97.46 & 79.06 & 90.15 & 47.74 & 47.00 & 89.68 & 57.99 & 94.42 & 93.27 & 72.14 & 79.84 & 59.16 & 74.33 & 53.64 & 48.00 & 67.46 & 55.30 & 57.01 & 70.92 & 60.95 & 75.66 & 70.24 \\
& 0.0870  & 97.58 & 80.22 & 90.50 & 47.41 & 47.20 & 90.25 & 58.40 & 94.54 & 93.58 & 75.20 & 83.35 & 57.80 & 75.77 & 54.76 & 52.58 & 69.71 & 56.64 & 57.73 & 72.09 & 62.75 & 76.34 & 71.33 \\
& 0.1279 & 97.71 & 81.28 & 91.35 & 49.49 & 52.74 & 91.10 & 58.25 & 94.82 & 94.16 & 73.89 & \textbf{88.41} & 70.80 & 78.11 & 58.23 & 58.27 & 72.40 & 59.67 & 61.75 & 74.68 & 66.16 & 78.67 & 74.06 \\

\midrule[0pt]
\parbox[t]{.5mm}{\multirow{4}{*}{\rotatebox[origin=c]{90}{AO (BPG)}}} & 0.0454 & 97.51 & 79.73 & 91.04 & 45.38 & 48.27 & 90.81 & \textbf{57.10} & 94.55 & \textbf{94.36} & \textbf{81.44} & 85.13 & \textbf{74.56} & 77.34 & 56.23 & 58.89 & 72.52 & 57.51 & 61.33 & 72.30 & 65.16 & 78.25 & 73.47 \\
& 0.0674 & 97.67 & 81.13 & \textbf{91.86} & \textbf{54.33} & 54.61 & 91.40 & 57.30 & 94.82 & 94.46 & \textbf{82.53} & 84.56 & 63.39 & 78.86 & \textbf{57.97} & 58.46 & 73.55 & 60.17 & 63.75 & 74.09 & 66.69 & 78.99 & 74.47\\
& 0.0870 & 97.42 & 80.18 & 91.89 & 53.25 & 51.30 & 91.65 & \textbf{59.39} & 94.65 & 94.64 & 78.62 & 85.76 & 69.54 & 79.71 & 59.47 & 59.62 & \textbf{74.63} & 61.01 & 63.92 & 75.82 & 67.74 & 79.02 & 74.87\\
& 0.1279 & 97.70 & 81.86 & 92.04 & 48.35 & 52.93 & 91.94 & \textbf{62.68} & 95.03 & 94.91 & \textbf{81.08} & 88.30 & 71.81 & 80.72 & \textbf{60.90} & \textbf{64.78} & 75.66 & 62.09 & 65.61 & 77.13 & 69.56 & 79.89 & 76.08 \\

\midrule[0pt]
\parbox[t]{.5mm}{\multirow{4}{*}{\rotatebox[origin=c]{90}{RO (BPG)}}} & 0.0454 & \textbf{97.78} & \textbf{81.69} & \textbf{91.52} & \textbf{50.05} & \textbf{52.45} & \textbf{91.24} & 56.78 & \textbf{94.97} & 94.18 & 78.88 & \textbf{86.13} & 73.22 & \textbf{77.88} & \textbf{56.31} & \textbf{62.19} & \textbf{73.15} & \textbf{59.39} & \textbf{62.81} & \textbf{74.47} & \textbf{66.60} & \textbf{79.07} & \textbf{74.48} \\
 & 0.0674 & \textbf{97.95} & \textbf{82.89} & 91.84 & 48.17 & \textbf{52.14} & \textbf{91.57} & 59.47 & \textbf{94.85} & \textbf{94.53} & 78.86 & \textbf{85.97} & \textbf{76.59} & \textbf{79.37} & \textbf{58.51} & \textbf{62.00} & \textbf{74.07} & \textbf{60.90} & \textbf{65.14} & \textbf{75.25} & \textbf{67.89} & \textbf{79.57} & \textbf{75.27} \\
 & 0.0870 & 97.71 & 81.73 & \textbf{92.13} & \textbf{52.62} & \textbf{54.76} & \textbf{91.75} & 56.93 & 94.82 & \textbf{94.81} & \textbf{83.52} & \textbf{88.36} & \textbf{78.94} & \textbf{80.10} & 59.45 & \textbf{62.72} & \textbf{74.63} & \textbf{62.07} & \textbf{64.93} &\textbf{76.44} & \textbf{68.62} & \textbf{80.67} & \textbf{76.23} \\
 & 0.1279 & 97.58 & 81.57 & \textbf{92.54} & \textbf{53.23} &\textbf{58.71} & \textbf{92.05} & 61.16 & 94.85 & \textbf{95.05} & 79.79 & 85.14 & 69.27 & \textbf{80.91} & 60.34 & \textbf{64.78} & \textbf{76.11} & \textbf{63.11} & \textbf{66.83} & \textbf{77.84} & \textbf{69.99} & 80.08 & 76.36 \\
\midrule[1pt]
{\setlength{\tabcolsep}{0pt} \begin{tabular}{c}{OO}\\(PNG)\end{tabular}} & 9.02 & 98.17 & 85.04 & 92.84 & 54.96 & 61.15 & 92.63 & 64.62 & 95.05 & 95.50 & 85.62 & 87.02 & 70.18 & 82.57 & 63.08 & 66.79 & 78.00 & 64.95 & 69.70 & 78.69 & 71.97 & 81.90 & 78.24 \\
\bottomrule
\end{tabular}}
\label{tab:cityscapes-per_class_miou}
\end{table*}
}

\newcommand{\tablehybridtrainingsets}{
\begin{table}
\centering
\caption{Segmentation performance of hybrid training sets compared against other configurations.}
\begin{tabular}{cccccccc}
\toprule
Compression method & CO & rO & o+c O & o+r O  \\ \midrule[1pt]
MSH at 0.0419 bpp & 69.55 & 74.07 & \textbf{75.96} & 73.99 \\
MSH at 0.0891 bpp & 72.54 & 75.69 & \textbf{77.08} & 75.88 \\
\midrule[0pt]
BPG at 0.0454 bpp & 71.49 & 72.32 & \textbf{74.42} & 73.98 \\
BPG at 0.0674 bpp & 73.03 & 74.01 & 75.87 & \textbf{76.12} \\
BPG at 0.0870 bpp & 73.54 & 75.15 & 75.98 &\textbf{76.67} \\
\bottomrule
\end{tabular}
\label{tab:hybridtrainingsets}

\vspace{0.5em} The original images form 12.5\% of all mixtures and the configuration oO results in a performance of 71.90. The IoU of the \textit{Train} class affects the mIoU dramatically affects BPG at 0.0454 bpp. When \textit{Train} class is excluded from mIoU, the values for o+c O and o+r O become 74.59 and \textbf{74.96}.

\end{table}
}

\newcommand{\tableperceptualindices}{
\begin{table}
\centering
\caption{Perceptual indices of different image sets.}
\resizebox{\linewidth}{!}{
\begin{tabular}{cccccc}
\toprule
& & &  
\multicolumn{3}{c}{Restoration type} \\
\multirow{2}{*}{\centering Compression method}& \multirow{2}{*}{Original}  & \multirow{2}{*}{Compressed} & \multirow{2}{*}{Adversarial} & \multicolumn{2}{c}{Non Adversarial} \\
& & & & RDN - P & RDN - M  \\
\midrule
MSH at 0.0419 bpp & 40.21 & 55.69 & 40.56 &  55.40 &  55.82\\
\midrule
BPG at 0.0674 bpp & 40.21 & 52.95 & 41.10 &  53.69 &  53.62 \\
\bottomrule
\end{tabular}}
\label{tab:perceptualindices}

\vspace{0.5em} RDN - P and RDN - M stand for RDN (PSNR) and RDN (MS-SSIM) respectively. The perceptual indices are calculated using the Blind image quality assessment method of HOSA~\cite{xu2016blind}.

\end{table}
}

\newcommand{\tablenewrdnresults}{
\begin{table}
\centering
\caption{Comparison of adversarial restoration against RDN in various metrics.}
\setlength{\tabcolsep}{4pt}
\resizebox{\columnwidth}{!}{
\begin{tabular}{cccccc}
\toprule
\multirow{3}{5em}{\centering Compression method} & \multirow{3}{5em}{\centering Evaluation metric} &  \multicolumn{4}{c}{\centering Restoration type}\\
&  & None & \multicolumn{2}{c}{Non adversarial} & Adversarial \\
& & (CO) & RDN - P & RDN - M & (RO) \\
\midrule[1pt]
\multirow{3}{5em}{ \centering MSH at 0.0419 bpp } & mIoU & 69.55 & 71.63 & 71.82 & \textbf{74.32} \\
&PSNR (dB) & 33.47 & \textbf{33.67} & 33.51 & 31.55 \\
&MS-SSIM (dB) & 13.43 & \textbf{13.66}& \textbf{13.66} & 11.87 \\
\midrule[0pt]

\multirow{3}{5em}{\centering MSH at 0.0613 bpp } & mIoU & 72.82 & 72.39 & 71.92 & \textbf{74.94} \\
&PSNR (dB) & 35.00 & \textbf{35.17} & 35.00 & 32.86 \\
&MS-SSIM (dB) & 14.90 & \textbf{15.11} & \textbf{15.10} & 13.08 \\
\midrule[0pt]

\multirow{3}{5em}{\centering BPG at 0.0454 bpp } & mIoU & 71.49 & 70.91 & 69.58 & \textbf{74.48} \\
&PSNR (dB) & 33.37 & \textbf{34.04} & 33.62 & 31.95  \\
&MS-SSIM (dB) & 12.96 & \textbf{13.58}  & \textbf{13.57} & 11.74 \\
\midrule[0pt]

\multirow{3}{5em}{\centering BPG at 0.0674 bpp } & mIoU & 73.03 & 73.26 & 73.96 & \textbf{75.27} \\
& PSNR (dB) & 34.81 & \textbf{35.56} & 35.19 & 33.40 \\
& MS-SSIM (dB) & 14.31 & \textbf{14.99} & \textbf{14.96} & 13.01 \\
\bottomrule
\end{tabular}
}

\vspace{0.5em} RDN - P and RDN - M stand for RDN (PSNR) and RDN (MS-SSIM).

\label{tab:rdn_results}
\end{table}
}

\newcommand{\tableefficiency}{
\begin{table}
\centering
\caption{Inference times for different approaches.}
\begin{tabular}{c cc}
\toprule
\multirow{2}{5em}{\centering Compression method} & Segment & Encode + Decode + Segment  \\
& OO / CO / RO & CC \\
\midrule
MSH & 0.64s & 0.10s + 0.13s + 0.64s = 0.87s \\
BPG & 0.64s & 1.10s + 0.15s + 0.64s = 1.89s \\
\bottomrule
\end{tabular}
\label{tab:efficiency}

\vspace{0.5em} Note that Encode and Decode times for MSH and BPG are measured on a GPU and a CPU respectively. Segmentation time is measured on a GPU. 
\end{table}
}

\newcommand{\figuredatacollection}{

\setlength{\fboxsep}{0pt}
\begin{figure*}[htbp]
	\centering
	{\setlength{\tabcolsep}{3pt}
	 \renewcommand{\arraystretch}{1.5}
	\begin{tabular}{ccc}
		\includegraphics[width=0.4\textwidth]{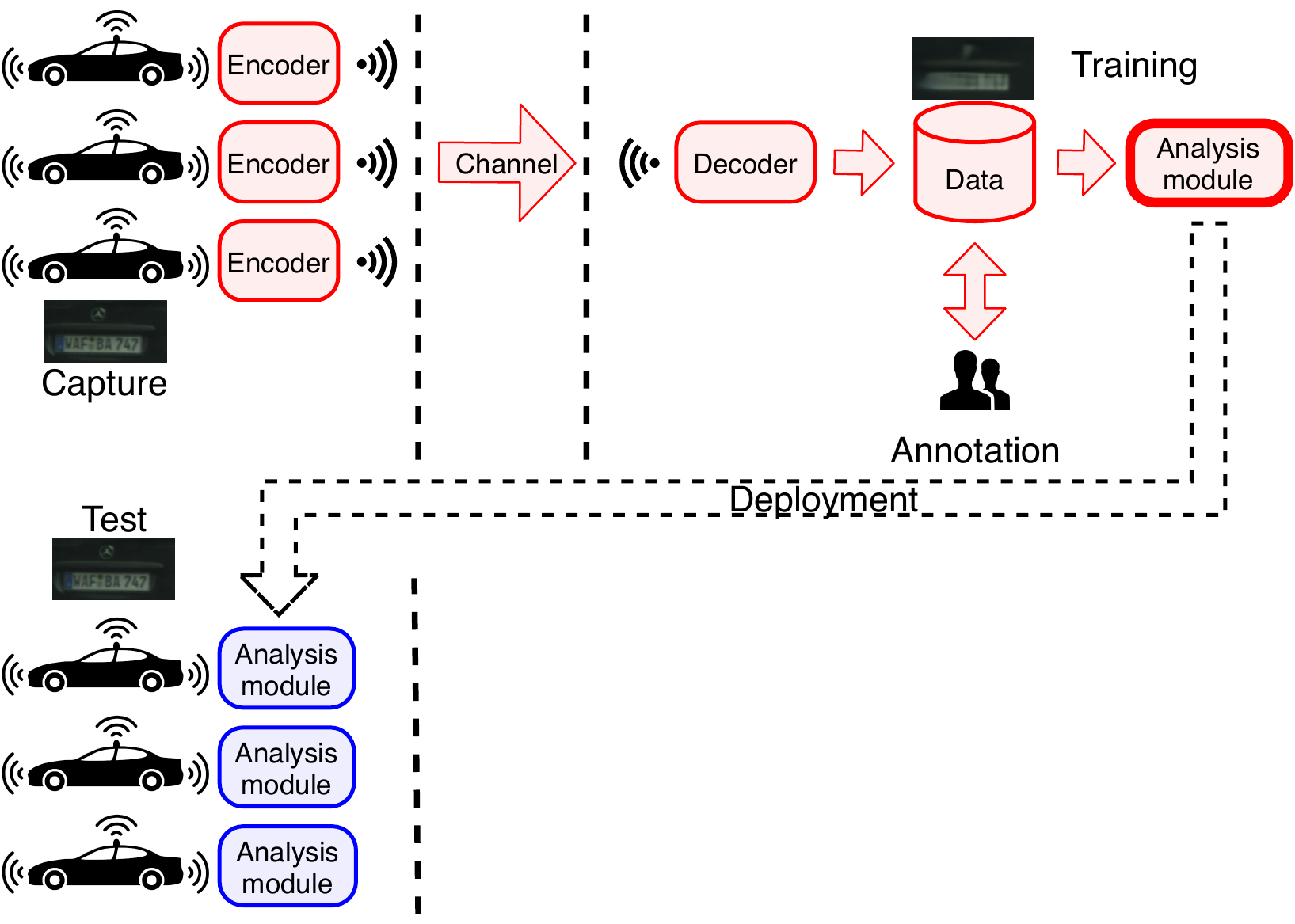} & {\setlength{\tabcolsep}{0pt}
			\renewcommand{\arraystretch}{0}
			\raisebox{\height}{\begin{tabular}{ccc}
			\includegraphics[width=0.1\textwidth]{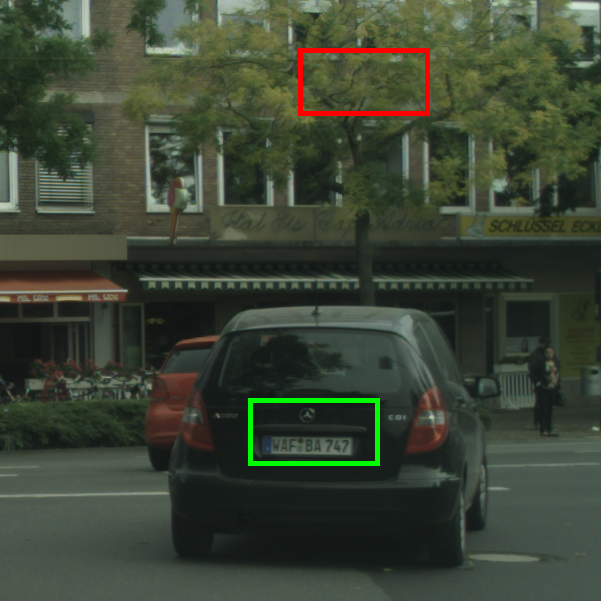} & \includegraphics[width=0.1\textwidth]{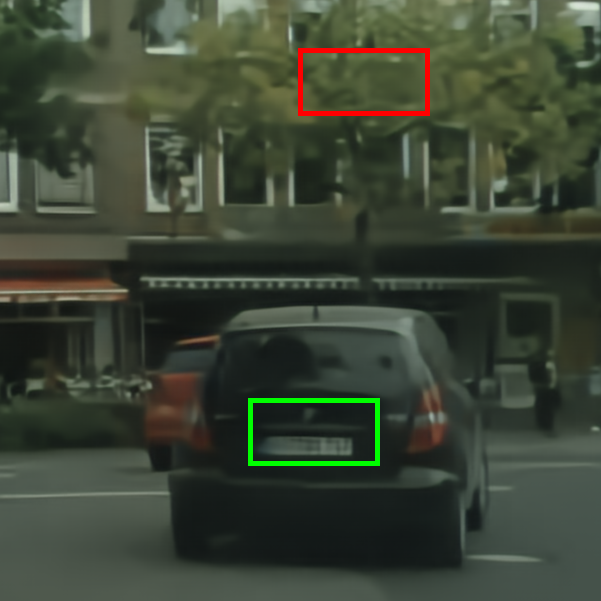} & \includegraphics[width=0.1\textwidth]{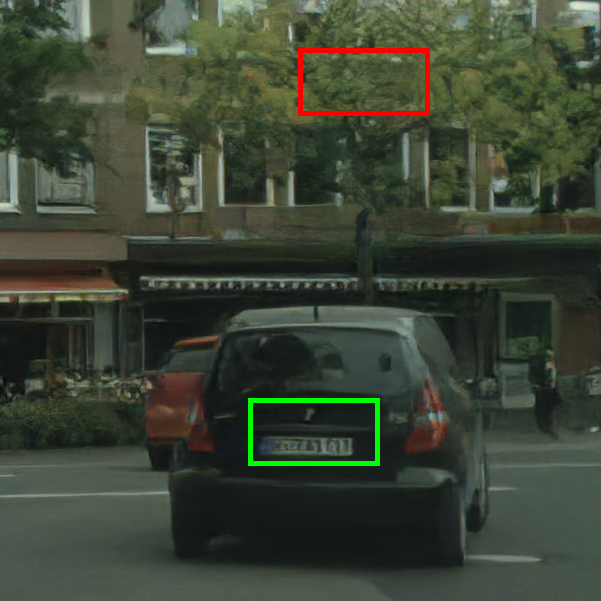}\\
			\fcolorbox{red}{white}{\includegraphics[width=0.1\textwidth]{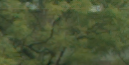}} & \fcolorbox{red}{white}{\includegraphics[width=0.1\textwidth]{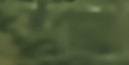}} & \fcolorbox{red}{white}{\includegraphics[width=0.1\textwidth]{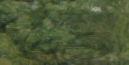}}\\
			\fcolorbox{green}{white}{\includegraphics[width=0.1\textwidth]{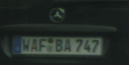}} & \fcolorbox{green}{white}{\includegraphics[width=0.1\textwidth]{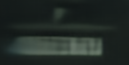}} & \fcolorbox{green}{white}{\includegraphics[width=0.1\textwidth]{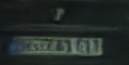}}\\			 \parbox{0.1\textwidth}{\centering\small Captured} & \parbox{0.1\textwidth}{\centering\small Compressed} & \parbox{0.1\textwidth}{\centering\small Restored}
		\end{tabular}} } & 
		\raisebox{0.\height}{\includegraphics[width=0.15\textwidth]{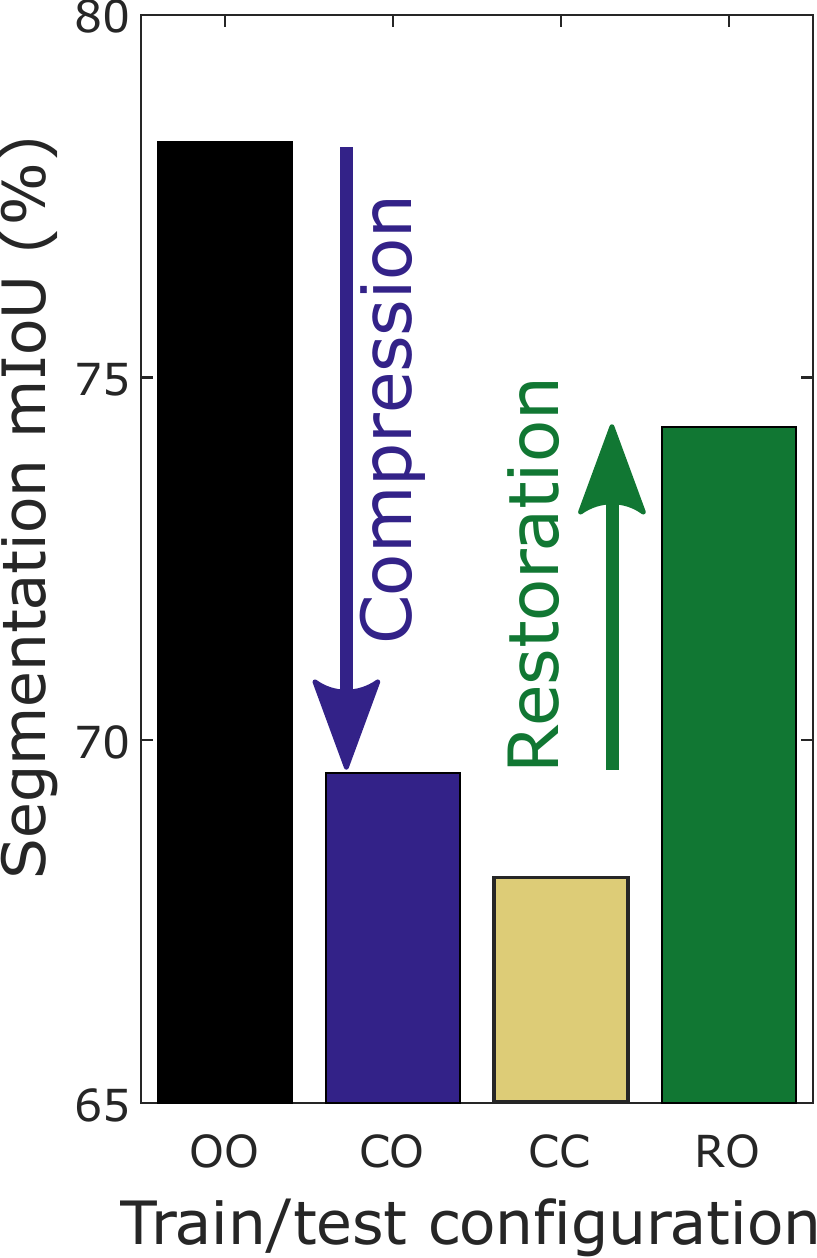}} \\
		(a) & (b) & (c)
	\end{tabular}
}
	\caption{Problem statement and proposed approach: (a) data collection using lossy compression makes training (top) and test data (bottom) different, (b) differences between test and training data are alleviated using adversarial restoration, and (c) drop in segmentation performance due to lossy compressed training data (CO/CC) and benefit from the proposed restoration method (RO).}
	\label{fig:feature_vis}
\end{figure*}
}

\newcommand{\figuredegradationexamples}{
	\begin{figure*}
		\centering
		\includegraphics[width=\textwidth]{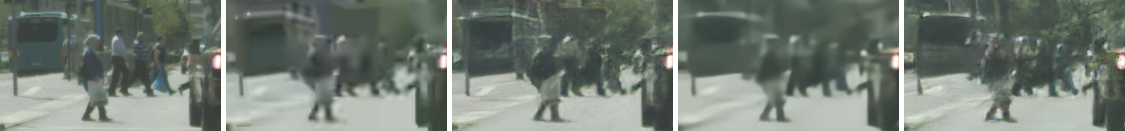}
		\includegraphics[width=\textwidth]{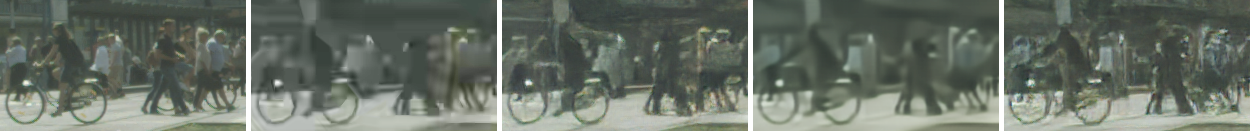}
		\includegraphics[width=\textwidth]{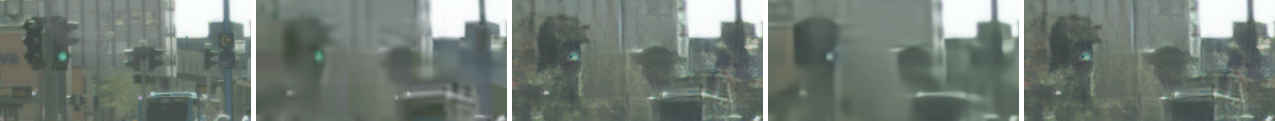}
		\caption{Effects of compression and restoration (from left to right): captured image, compressed (BPG), restored (BPG), compressed (MSH), restored (MSH). The brightness of the image crops have been slightly enhanced to improve visibility.}
		\label{fig:degradationexamples}
	\end{figure*}
}

\newcommand{\figuremcc}{
	\begin{figure}
		\centering
		\includegraphics[width=0.99\columnwidth]{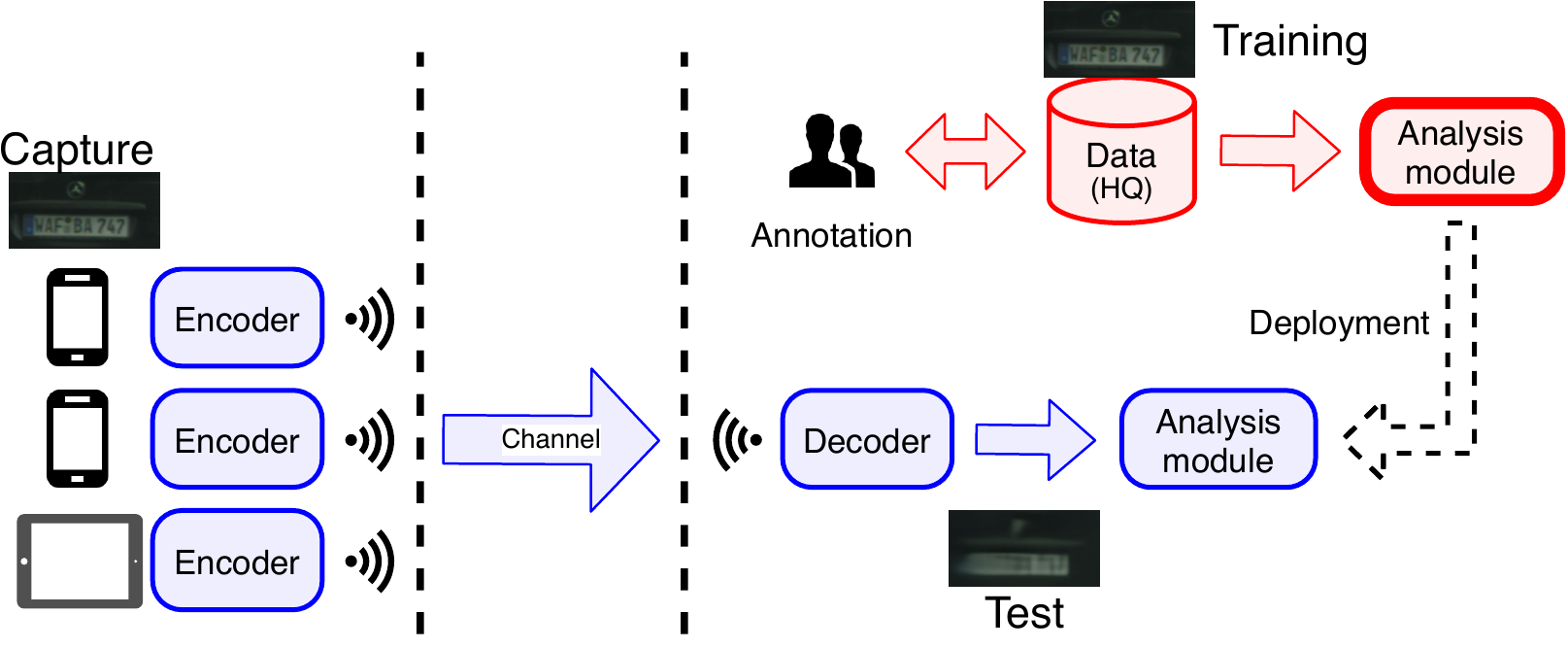}
		\caption{Example of OC configuration: mobile cloud computing with inference on compressed test images, and high quality training images.}
			\label{fig:mcc}
\end{figure}}

\newcommand{\figureaadaptationmethods}{
    \begin{figure}
		\centering
		{
	        \includegraphics[width=0.99\columnwidth]{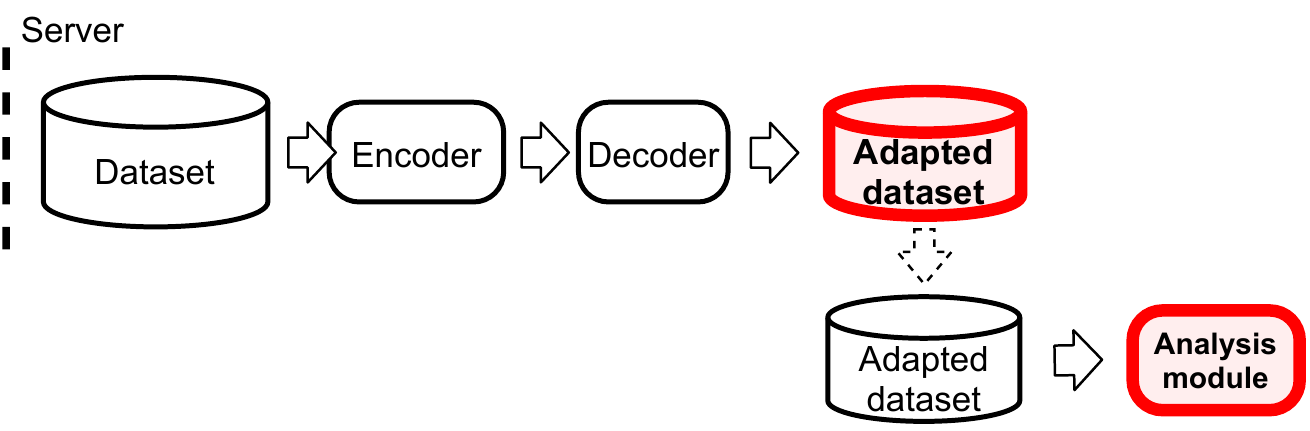}\\
			(a)\\
	        \includegraphics[width=0.99\columnwidth]{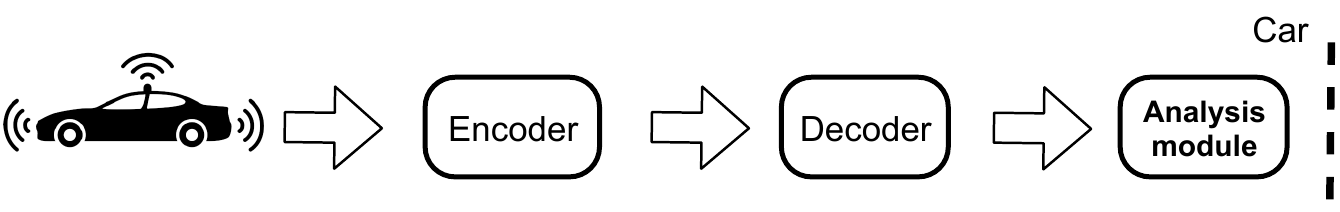}\\
			(b)\\
	        \includegraphics[width=0.99\columnwidth]{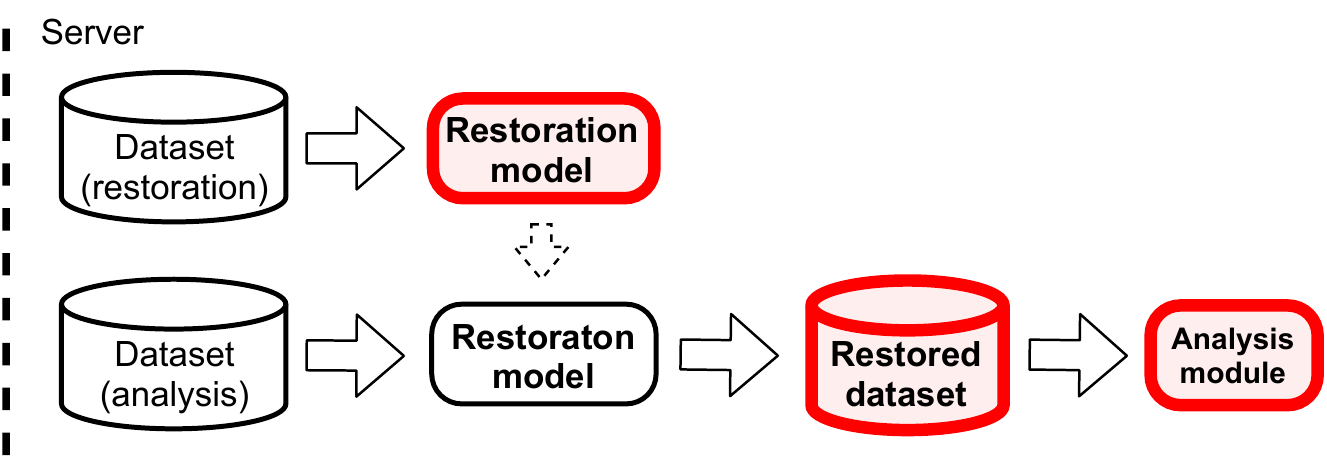}\\
	        (c)
		}
	\caption{Adaptation strategies: (a) compression before training, (b) compression before inference, and (c) dataset restoration.}
	\label{fig:adaptation_methods}
	\end{figure}
}

\newcommand{\figureilluspercepdist}{
\begin{figure}
    \centering
    \includegraphics[width=.73\columnwidth]{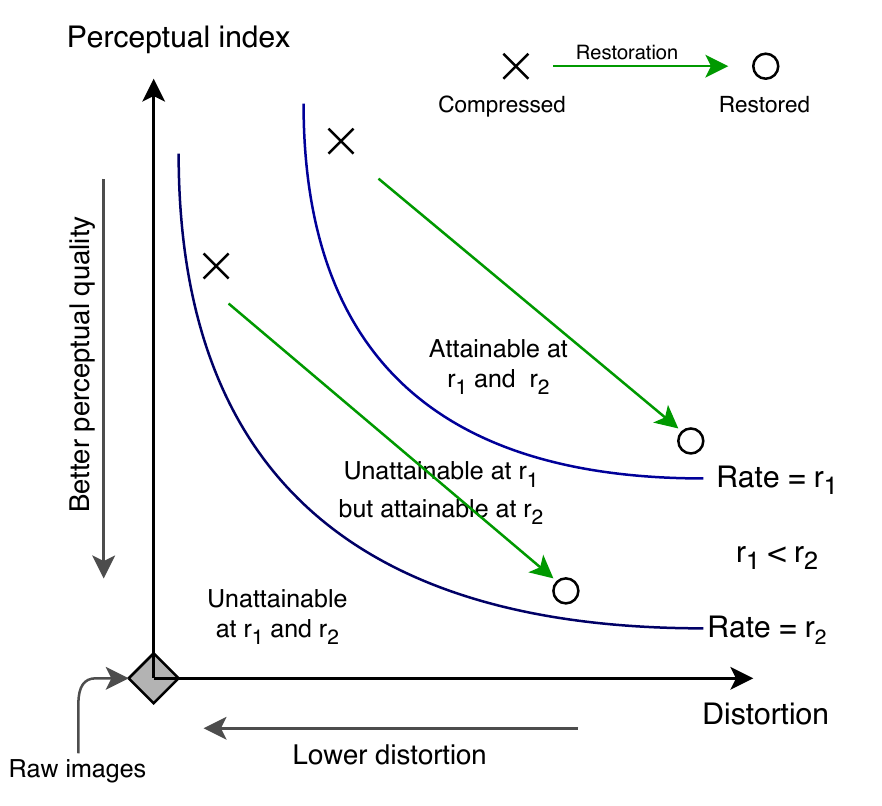}
    \caption{Illustration of the perception-distortion tradeoff~\cite{blau2018perception,blau2019rethinking}. Restoration process shifts the compressed images from a point of low distortion and low perceptual quality to a point of higher distortion but higher perceptual quality in the perception-distortion plane.}
    \label{fig:illus_percep_dist}
\end{figure}
}

\newcommand{\figurecityscapesrdcurvesnew}{
\begin{figure}
    \centering
    \includegraphics[width=0.99\columnwidth]{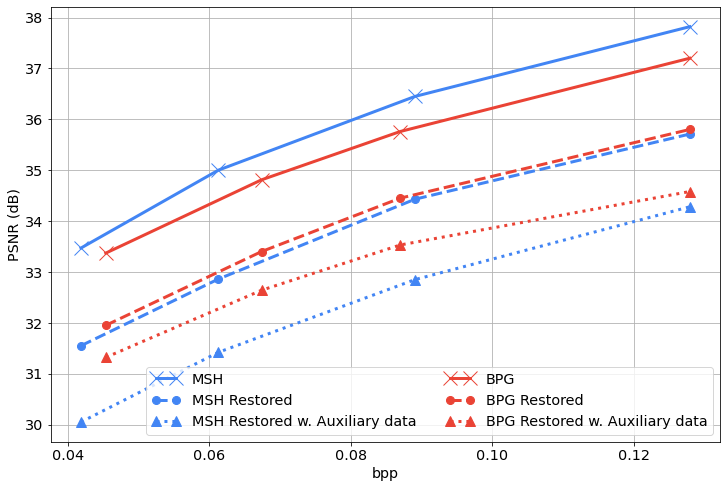}
    \caption{Rate-distortion curves for Cityscapes for BPG and MSH, and with and without adversarial image restoration.}
    \label{fig:cityscapes_rd_curves}
\end{figure}
}

\newcommand{\figurecityscapessegmentationperformance}{
\begin{figure}
    \centering
    \includegraphics[width=0.99\columnwidth]{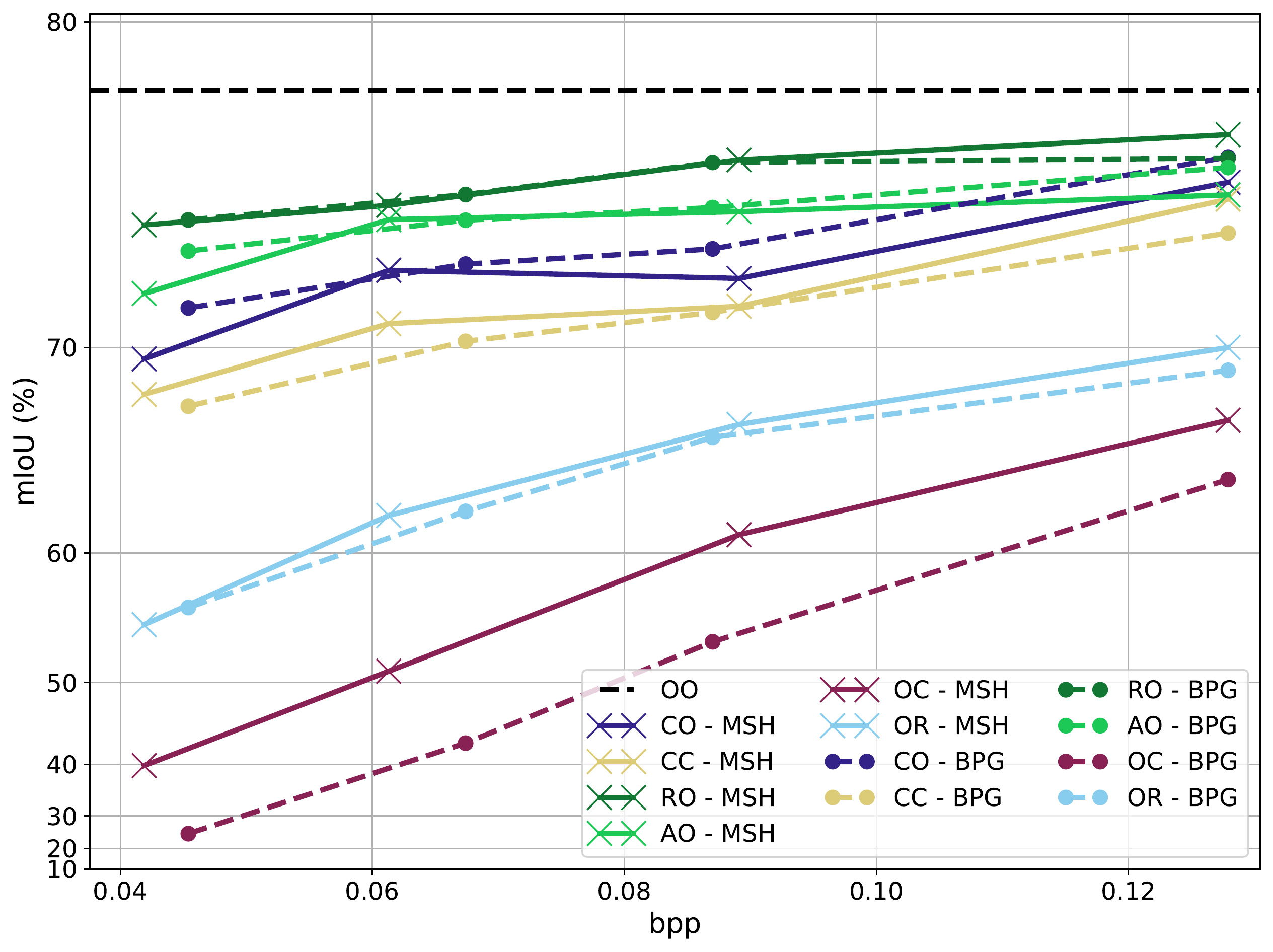}
    \caption{Segmentation performance on Cityscapes for different training/test configurations. }
    \label{fig:cityscapes-segmentation-performance}
\end{figure}
}

\newcommand{\figurerestorationdatarequirednew}{
\begin{figure}
    \centering
    \includegraphics[width=0.99\columnwidth]{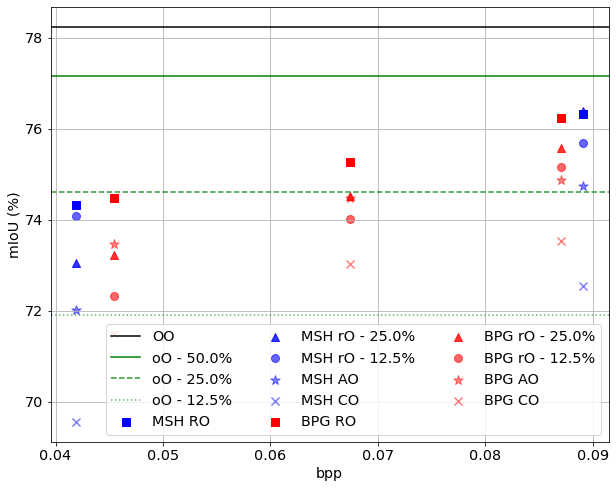}
    \caption{Performance of models with different configurations obtained by varying the amount of privileged data.}
    \label{fig:restoration_data-required}
\end{figure}
}

\newcommand{\figurecostperformance}{
	\begin{figure*}
		\centering
		\includegraphics[width=0.99\textwidth]{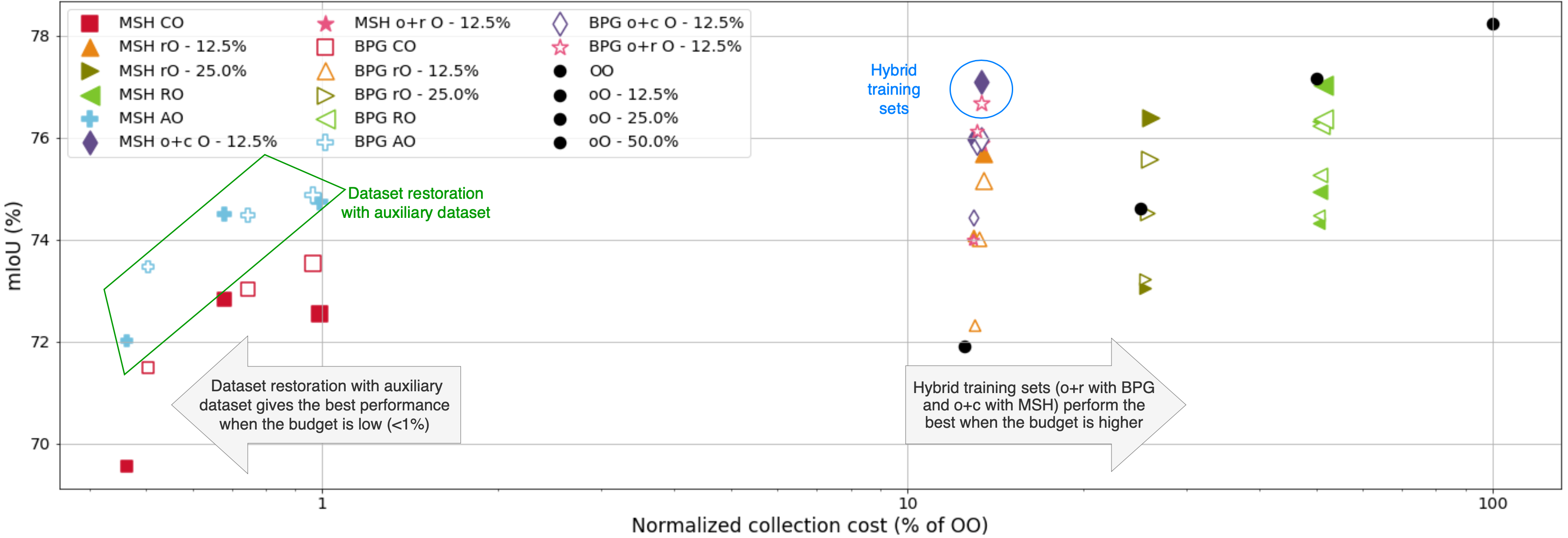}
		\caption{Segmentation performance of each configuration against cost of collecting data. Marker size indicates rate.}
		\label{fig:costperformance}
	\end{figure*}
}

\newcommand{\figureadversarialvsnonadversarial}{
	\begin{figure}
		\centering
		\includegraphics[width=0.99\columnwidth]{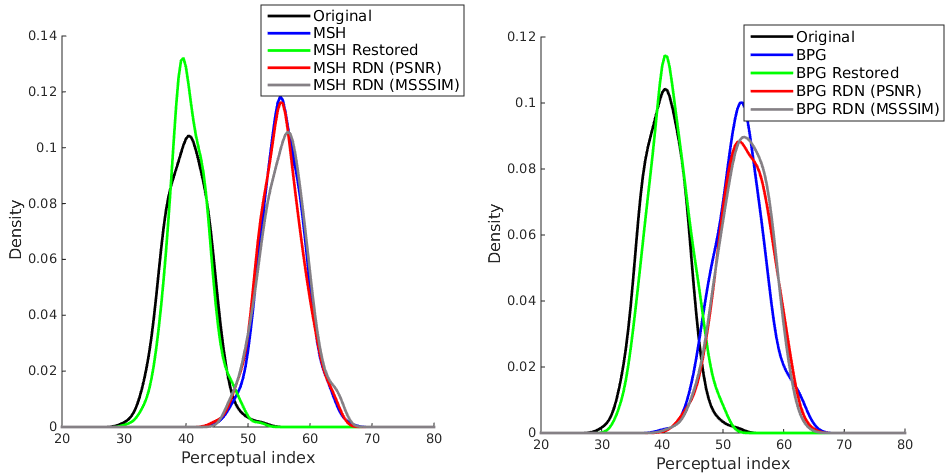}\\
	\caption{Adversarial dataset restoration vs non-adversarial: While Table~\ref{tab:perceptualindices} describes the average perceptual index of the original, compressed and restored image sets, this figure shows the distribution of the perceptual index. Dataset restoration with adversarial image restoration can recover the distribution of the perceptual indices of the original images, while non-adversarial cannot.}
		\label{fig:adversarial_vs_nonadversarial}
	\end{figure}
}

\newcommand{\figureimagesfeatcorr}{
\begin{figure}
    \centering
    \includegraphics[width=0.96\columnwidth]{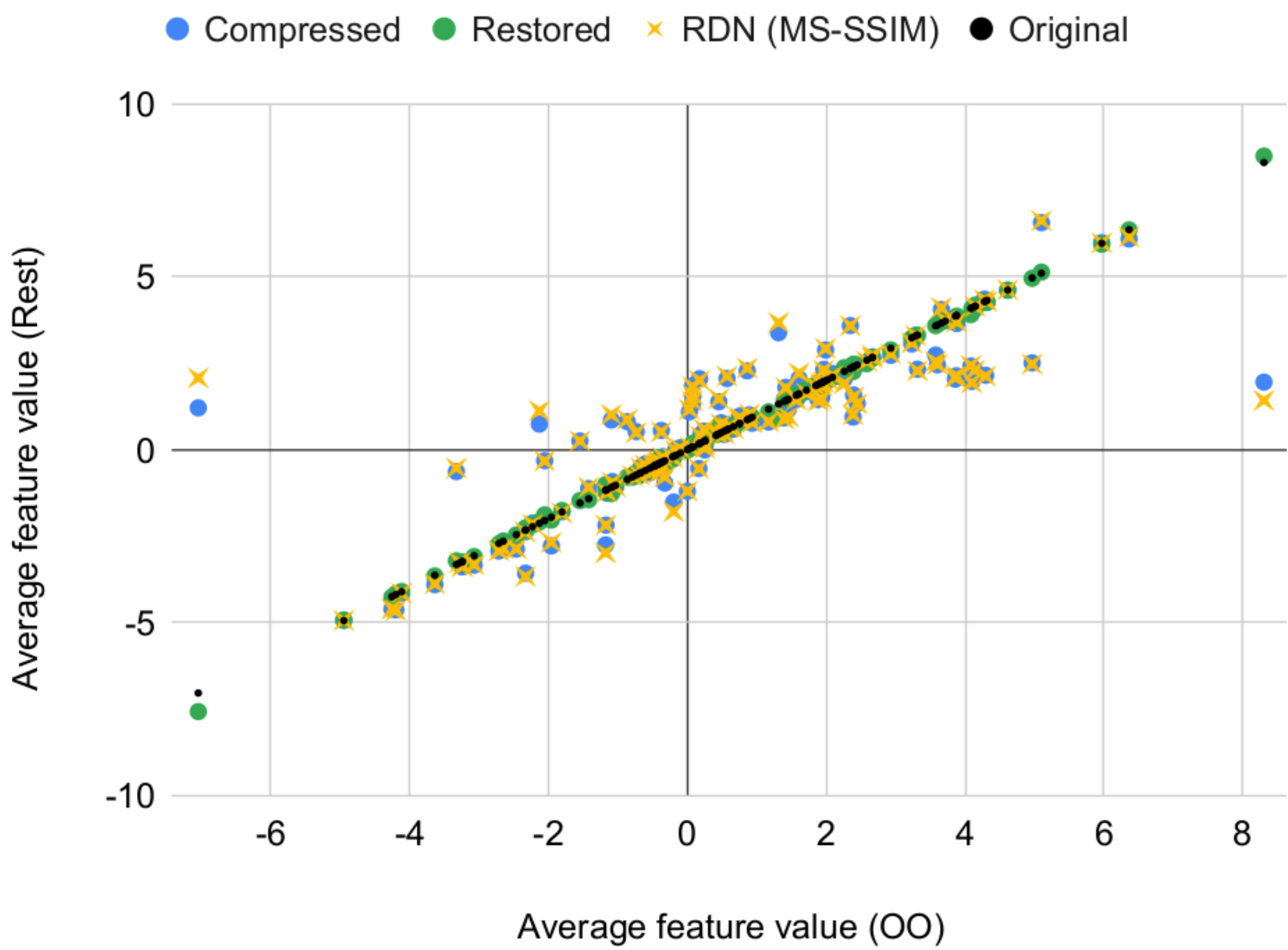}
    \caption{Feature correlation between segmentation features (at a shallow layer) of the segmentation model trained on raw images. Dataset restoration with adversarial image restoration can recover the distribution of perceptual index and segmentation features of the optimal case (OO), while non-adversarial cannot.}
    \label{fig:images_feat_corr}
\end{figure}
}

\newcommand{\figureaidsegmentationperformance}{
\begin{figure}
    \centering
    \includegraphics[width=0.99\columnwidth]{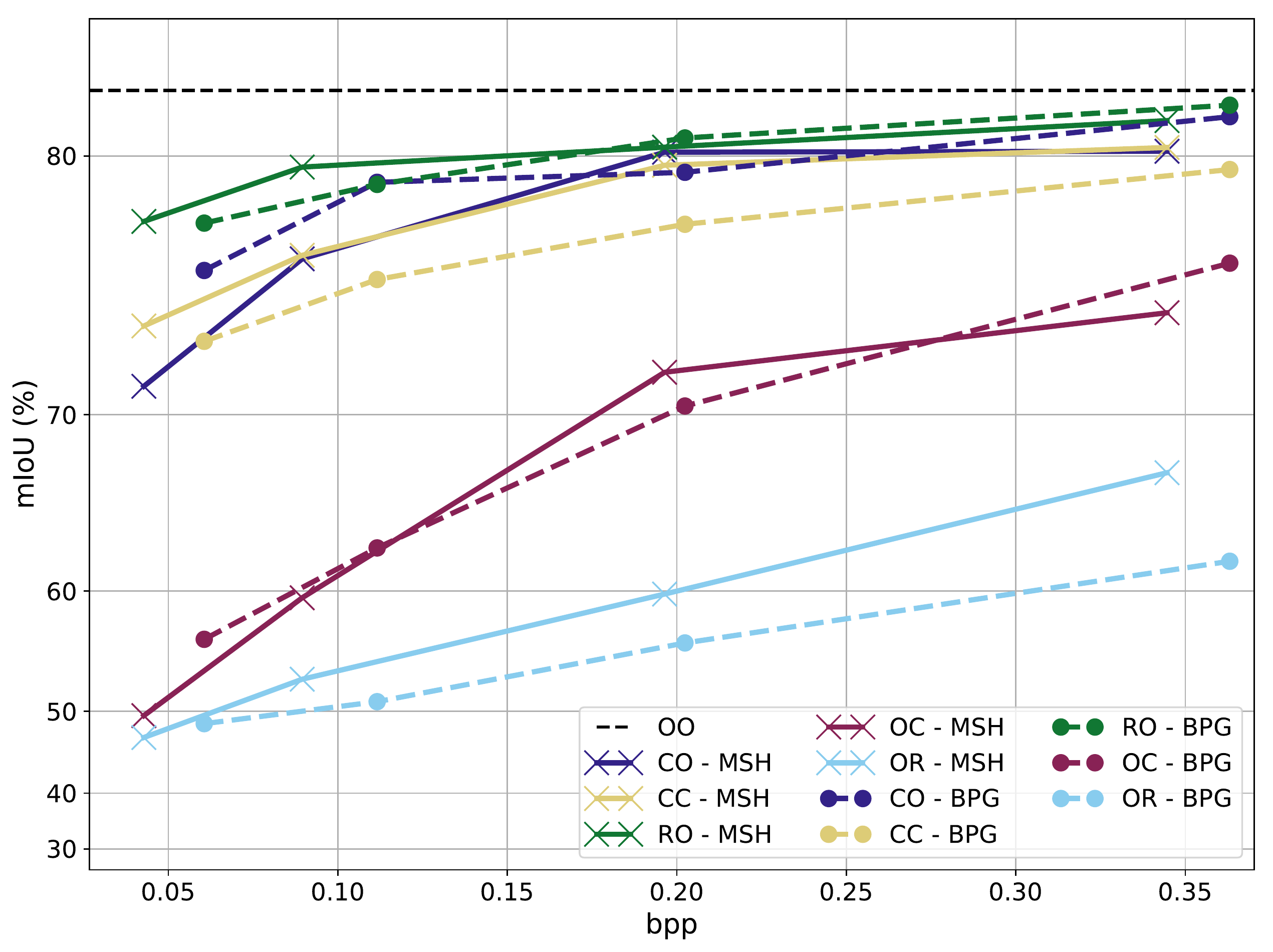}
    \caption{Segmentation performance on INRIA Aerial Images Dataset for different training/test configurations.}
    \label{fig:aid-segmentation-performance}
\end{figure}
}

\newcommand{\figureaidexample}{
	\begin{figure*}
		\centering
		\includegraphics[width=0.75\textwidth]{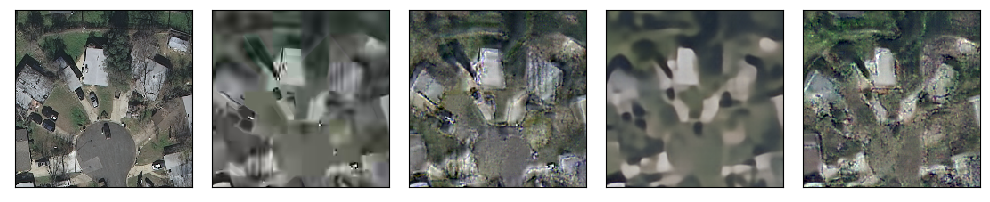}
		\includegraphics[width=0.50\textwidth]{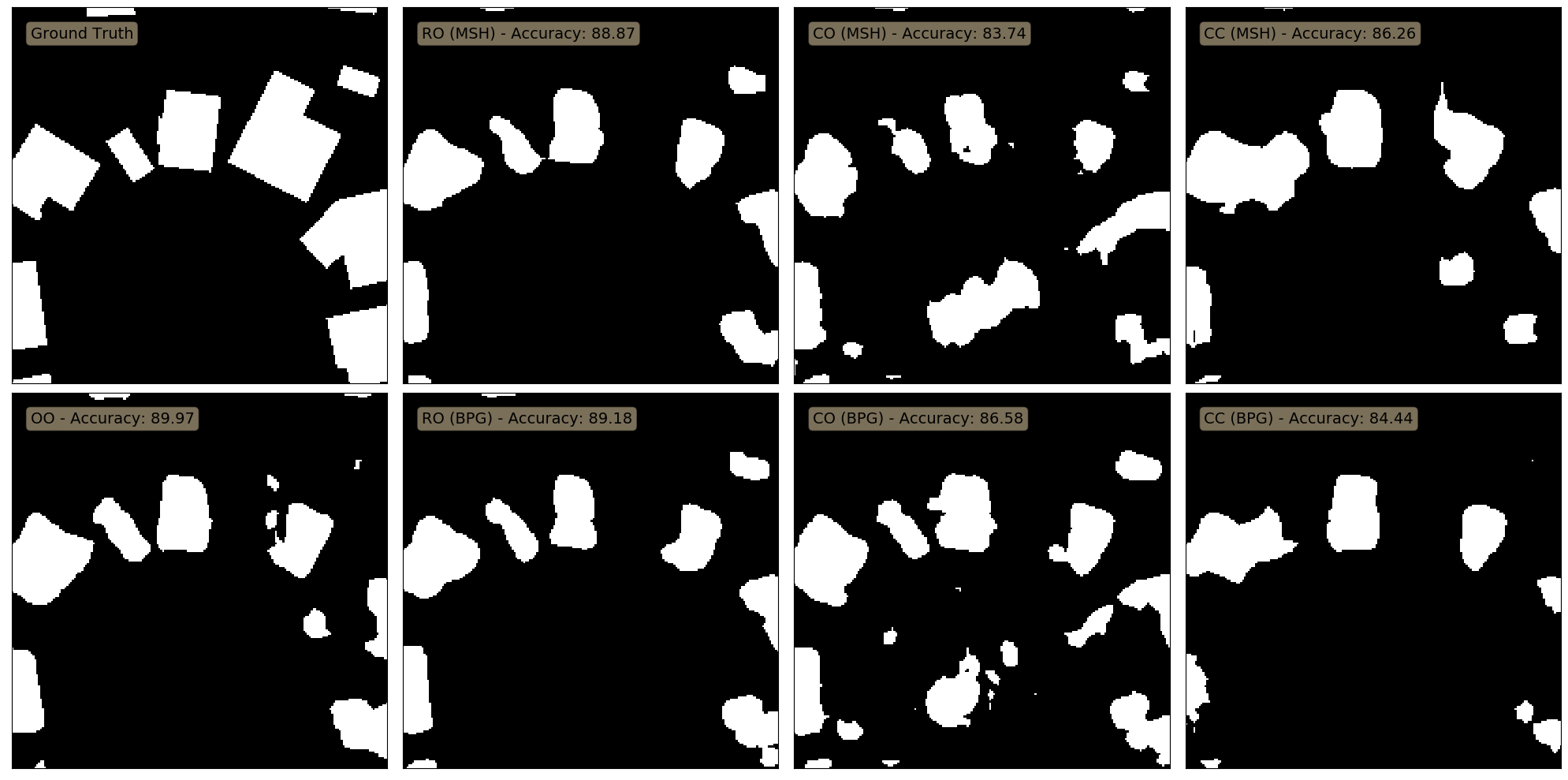}
		\caption{Top: (from left to right) - captured image, compressed (BPG), restored (BPG), compressed (MSH), restored (MSH). Bottom: Prediction map and accuracy score of segmentation models with different configurations.}
		\label{fig:aidexample}
	\end{figure*}
}

\newcommand{\figuresddsegmentationperformance}{
\begin{figure}
    \centering
    \includegraphics[width=0.99\columnwidth]{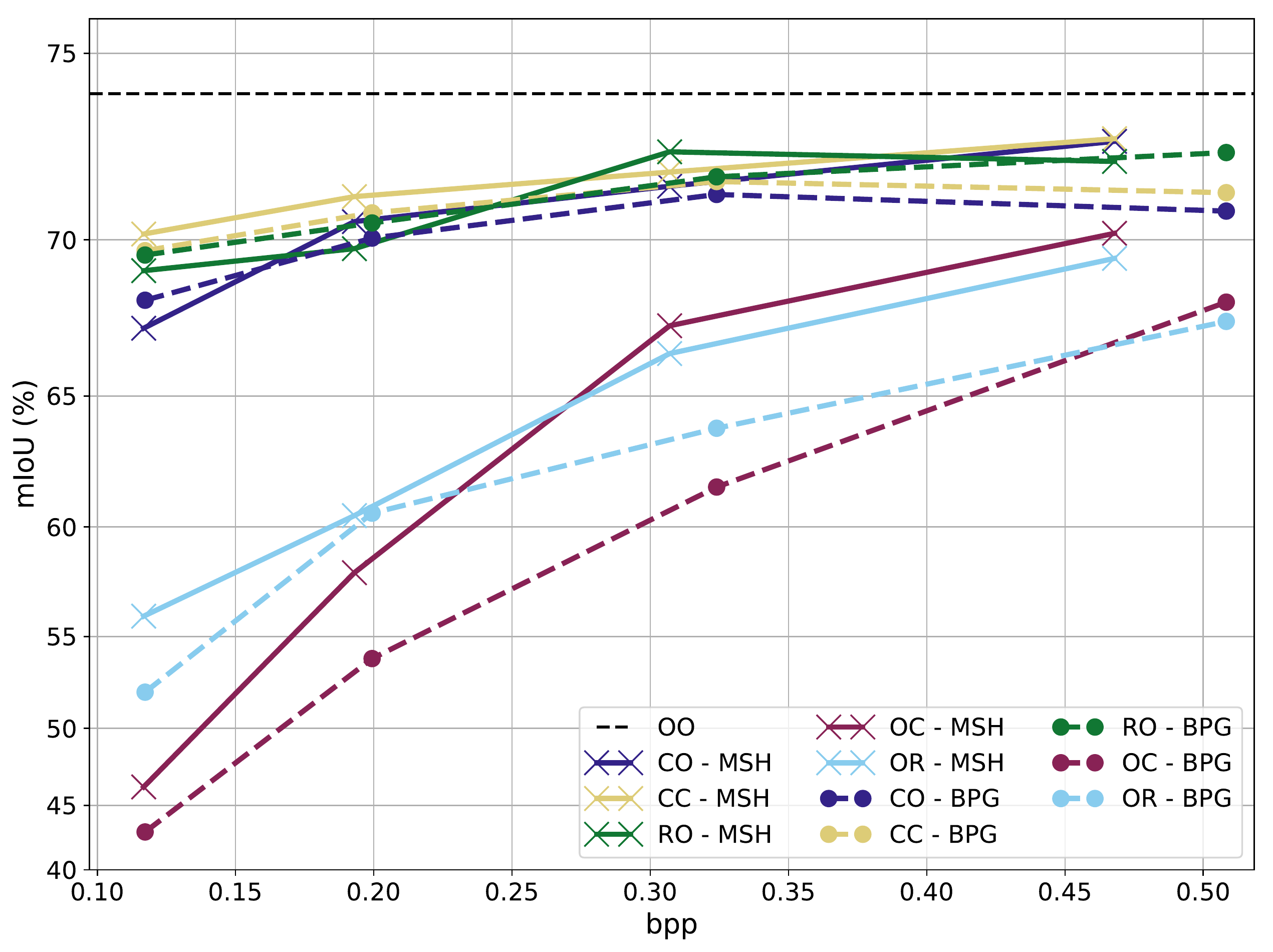}
    \caption{Segmentation performance on Semantic Drones Dataset for different training/test configurations.}
    \label{fig:sdd-segmentation-performance}
\end{figure}
}

\newcommand{\figuremsharch}{
\begin{figure}
    \centering
    \includegraphics[width=0.98\columnwidth]{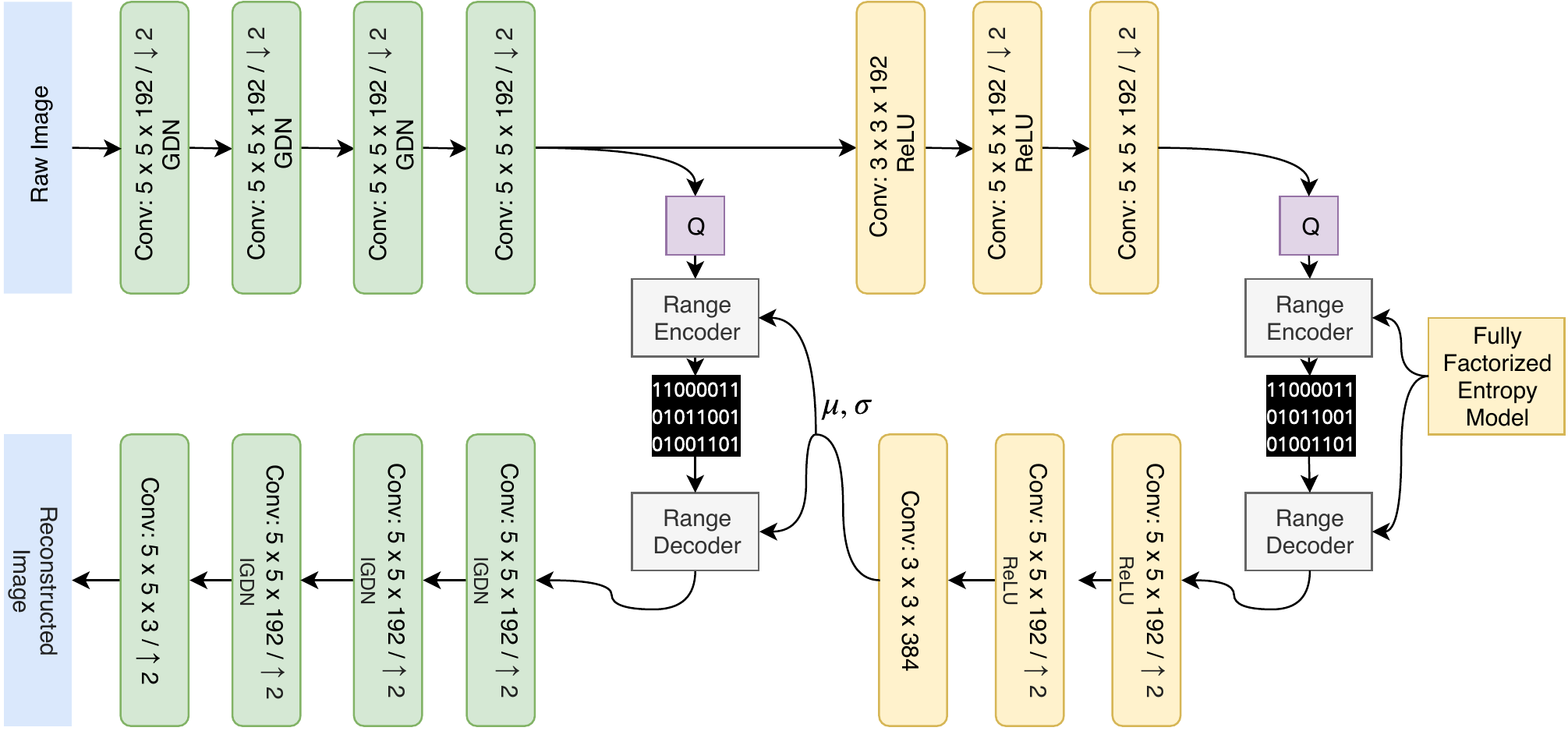}
    \caption{MSH Architecture: Refer to \cite{minnen2018joint} for details on Fully Factorized Entropy Model and Range Encoder-Decoder.}
    \label{fig:msh_arch}
\end{figure}
}

\begin{document}
	\title{Distributed Learning and Inference with Compressed Images}
	\author{
		Sudeep~Katakol,
		Basem~Elbarashy,
		Luis~Herranz,
		Joost~van~de~Weijer,
		and~Antonio~M.~L\'opez,~\IEEEmembership{Member,~IEEE}%
		
		\IEEEcompsocitemizethanks{
			\IEEEcompsocthanksitem S. Katakol is with the Department
			of Computer Science \& Information Systems and the Department of Mathematics, BITS Pilani, KK Birla Goa Campus, India, 403726. This work was done during an internship at Computer Vision Center, Barcelona. \protect\\
			E-mail: skatakol@cvc.uab.es
			
			\IEEEcompsocthanksitem B. Elbarashy, L. Herranz, J. van de Weijer and A. M. L\'opez are with the Computer Vision Center, Universitat Aut\`onoma de Barcelona, 08193 Bellaterra, Spain. J. van de Weijer and A. M. L\'opez are also associated with the Computer Science Dept. at Universitat Aut\`onoma de Barcelona.
		}%
	}
	
	\IEEEtitleabstractindextext{%
		\begin{abstract}
			Modern computer vision requires processing large amounts of data, both while training the model and/or during inference, once the model is deployed. Scenarios where images are captured and processed in physically separated locations are increasingly common (e.g. autonomous vehicles, cloud computing, smartphones). In addition, many devices suffer from limited resources to store or transmit data (e.g. storage space, channel capacity). In these scenarios, lossy image compression plays a crucial role to effectively increase the number of images collected under such constraints. However, lossy compression entails some undesired degradation of the data that may harm the performance of the downstream analysis task at hand, since important semantic information may be lost in the process. Moreover, we may only have compressed images at training time but are able to use original images at inference time (\textit{i.e.} test), or vice versa, and in such a case, the downstream model suffers from covariate shift. In this paper, we analyze this phenomenon, with a special focus on vision-based perception for autonomous driving as a paradigmatic scenario. We see that loss of semantic information and covariate shift do indeed exist, resulting in a drop in performance that depends on the compression rate. In order to address the problem, we propose dataset restoration, based on image restoration with generative adversarial networks (GANs). Our method is agnostic to both the particular image compression method and the downstream task; and has the advantage of not adding additional cost to the deployed models, which is particularly important in resource-limited devices. The presented experiments focus on semantic segmentation as a challenging use case, cover a broad range of compression rates and diverse datasets, and show how our method is able to significantly alleviate the negative effects of compression on the downstream visual task.
		\end{abstract}
		
		}
	
	\maketitle
	
	\IEEEdisplaynontitleabstractindextext
    \IEEEpeerreviewmaketitle

	\section{Introduction}
	\label{sec:introduction}
	
	\IEEEPARstart{M}{odern} intelligent devices such as smartphones, autonomous vehicles and robots are equipped with high-quality cameras and powerful deep neural networks that enable advanced on-board visual analysis and understanding. These large models are trained with a large amount of data and require powerful hardware resources (e.g. GPUs). These models also require days or even weeks to train, which is not possible in resource-limited devices. Thus, training is often performed in a centralized server, which also allows using data captured by multiple devices to train better models (e.g. a fleet of autonomous cars). In this case, training and testing take place in two physically separated locations, \textit{i.e.} server and device, respectively. In other cases, such as in mobile cloud computing, the data is captured by the device, while the inference takes place in a server.
	
	One important requirement in these scenarios is that, at some point, the visual data needs to be transmitted from the device to the server. Fig.~\ref{fig:feature_vis}a shows an archetypal scenario of autonomous driving, where each vehicle of the fleet captures and encodes data and transmits it to the server. The server decodes the data and uses it for training the analysis models. The trained models are then deployed to autonomous vehicles, where they perform inference. Captured data often requires to be annotated by humans in order to train supervised models, which adds to the reasons to process the data in a server.
	
	\figuredatacollection
	
	The captured data can be stored on-board in a storage device and physically delivered to the server, or directly transmitted through a communication channel. In either case, storage space or channel capacity are constraints that condition the amount of collected samples in practice, and effective collection requires data compression to exploit the limited storage and communication resources efficiently.
	
	The amount of data captured (possibly from multiple cameras) can be enormous, requiring high compression rates with lossy compression. However, this entails a certain degradation in the images, which depends on the bitstream rate (the lower the rate, the higher the degradation). In this paper, we study the impact of such degradation on the performance of the downstream analysis task. At times, the degradation affects only one of the training and test data. For instance, in Fig~\ref{fig:feature_vis}a, training data is degraded while test data (on-board) can be accessed without degradation. 
	
	When training data is compressed and test is not (or vice versa), a first effect we observe is \textit{covariate shift} (\textit{i.e.} the training and test data are sampled from different distributions). For instance, the first column of Fig.~\ref{fig:feature_vis}b represents the original captured images, while the second represents the compressed images (\textit{i.e.} reconstructed\footnote{When referring to data used in the downstream tasks, \textit{compressed} images will implicitly refer to the reconstructed images after the compression decoder.}). A clear difference in terms of lack of details and blurred textures is observed, which causes covariate shift (e.g. original for test, compressed for training in the example of Fig~\ref{fig:feature_vis}a). A possible solution to this problem is compressing both training and test data at the same rate. For the autonomous driving scenario of Fig.~\ref{fig:feature_vis}a, this would mean deploying an image compressor in the car (including encoder and decoder) and performing inference on the reconstructed images. While this approach alleviates the covariate shift, it is not always effective and also increases the computational cost in the on-board system. 
	
	The degradation caused by lossy compression not only induces covariate shift, but can also harm the performance of the downstream task through the means of  \textit{semantic information loss}. Here, semantic information refers to the information that is relevant to solve a particular downstream task and it can be lost during the process of compression. Semantic information is task-dependent and its loss is typically irreversible. For example, the actual plate number WAF BA 747 in the second column of Fig.~\ref{fig:feature_vis}b is lost in the process of compression, and cannot be recovered. However, if the task is car detection, the actual plate number is not necessarily relevant semantic information.
	
	In this paper, we study the effect of compression on downstream analysis tasks (focusing on semantic segmentation) under different configurations, which in turn can be related to real scenarios. We observe that both covariate shift and semantic information loss indeed result in a performance drop (see Fig.~\ref{fig:feature_vis}c\footnote{Segmentation performance is measured as the mean Intersection over Union (mIoU), which is the ratio between the correctly predicted pixels and union of predicted and ground truth pixels, averaged across every category.}) compared to training and test with original images (configuration OO). The performance depends on the compression rate and the particular training/test configuration. For instance, in the configuration of the autonomous driving scenario of Fig.~\ref{fig:feature_vis}a, compressing the test data prior to inference (we refer to this approach as \textit{compression before inference}, and corresponds to the training/test configuration compressed-compressed, or CC for short) degrades the performance more than using the original data (configuration CO), showing that it is preferable to keep the test data more semantically informative than correcting the covariate shift.
	
	The previous result also motivates us to explore whether there exists a solution that improves over the baseline CO and CC configurations. As a result, we propose \textit{dataset restoration}, an effective approach based on image restoration using generative adversarial networks (GANs)~\cite{goodfellow2014generative}. Dataset restoration is applied to the images in the training set without modifying the test images, effectively alleviating the covariate shift, while keeping the test data semantically informative. In this case, we show that the configuration restored-original (RO) does improve performance over the baselines (see Fig.~\ref{fig:feature_vis}c). An additional advantage is that there is no computational cost penalty nor additional hardware or software requirements in the deployed on-board system (in contrast to compressing the test data). Note also that our approach is generic and independent of the particular compression (deep or conventional) used to compress the images.
	
	Adversarial restoration decreases the covariate shift by hallucinating texture patterns that resemble those lost during compression while removing compression artifacts, both of which contribute to the covariate shift. The distribution of restored images is closer to the distribution of original images and thus the covariate shift is lower. Fig.~\ref{fig:feature_vis}b shows an example where the trees have lost their texture and appear essentially as blurred green areas. A segmentator trained with these images will expect trees to have this appearance, but during test they appear with the original texture and details of leaves and branches, which leads to poor performance. The restored image has textures that resemble real trees and contains less compression artifacts, which makes its distribution closer to that of the actual test images, contributing to a significant improvement in downstream performance (see Fig~\ref{fig:feature_vis}c). Note that adversarial restoration cannot recover certain semantic information. This example also illustrates the effect on semantic information. The license plate appears completely blurred due to compression. Note that adversarial restoration can recover the texture of digits (or even hallucinate random digits), which can be useful to improve car segmentation, but the original plate number is lost (\textit{i.e.} semantic information), which makes it impossible to perform license plate recognition at that compression rate.

	In summary, our contributions are as follows:
	\begin{itemize}
		\item Systematic analysis of training/test configurations with compression and relation of downstream performance with rate, semantic information loss and covariate shift.
		\item Dataset restoration, a principled method based on our theoretical analysis, to improve downstream performance in on-board analysis scenarios. This method is task-agnostic and can be used alongside multiple image compression methods. It also does not increase the inference time and memory requirements of the downstream model.
	\end{itemize}

	\section{Related Work}
	\label{sec:related_work}
	
	\subsection{Lossy compression}
	A fundamental problem in digital communication is the transmission of data as binary streams (\textit{bitstreams}) under limited capacity channels~\cite{shannon1948mathematical,ct91}, a problem addressed by data compression. Often, practical compression ratios are achievable only with lossy compression, \textit{i.e.} a certain loss with respect to the original data is tolerated. Traditional lossy compression algorithms for images typically use a DCT or a wavelet transform to transform the image into a compact representation, which is simplified further to achieve the desired bitrate. Examples of lossy image compression algorithms are JPEG~\cite{wallace1992jpeg}, JPEG 2000~\cite{skodras2001jpeg,taubman2012jpeg2000}, and BPG~\cite{bellard2017bpg}. BPG is the current state-of-the-art and is based on tools from the HEVC video coding standard~\cite{sullivan2012overview}. 
	
	Recently, deep image compression~\cite{balle2016end,balle2018variational,toderici2015variable,theis2017lossy,minnen2018joint,yang2020variable} has emerged as a powerful alternative to the traditional algorithms. These methods also use a transformation based approach like the traditional methods, but use deep neural networks to parameterize the transformation~\cite{balle2016end}. The parameters of the networks are learned by optimizing for a particular rate-distortion tradeoff on a chosen dataset. Mean Scale Hyperprior (MSH)~\cite{minnen2018joint}, a deep image compression method based on variational autoencoders and BPG are used as representative methods of deep learning based and traditional image compression respectively.
	
	\subsection{Visual degradation and deep learning}
	
	A loss in the quality of images can occur through many factors including blur, noise, downsampling and compression. Researchers have reported a drop in task performance of convolutional neural networks (CNN) models when such degradations are present in the test images~\cite{dodge2016understanding,hendrycks2019benchmarking,roy2018effects}. Further, numerous methods have been proposed to make these CNN models robust to degradations~\cite{hendrycks2019benchmarking,borkar2019deepcorrect,ghosh2018robustness}. 
	These approaches include forcing adversarial robustness during training~\cite{hendrycks2019benchmarking}, modifying and retraining the network~\cite{borkar2019deepcorrect}, and using an additional network altogether~\cite{ghosh2018robustness}.
	
	While the aforementioned works target robustness across degradations, there have been studies focusing exclusively on compression as well. These include~\cite{torfason2018towards} (on the deep compression method~\cite{theis2017lossy}),~\cite{ehrlich2019deep} (JPEG) and~\cite{lohdefink2019gan} (both deep~\cite{agustsson2019generative} and JPEG). Unlike the previous methods, these works use the compressed images (in some form) for training the deep models and thus obtain a better performance on compressed images. Moreover,~\cite{torfason2018towards} and~\cite{ehrlich2019deep} encode the images using the compressors and the deep networks are trained to predict the task output using the encoded representation directly, resulting in faster inference. 
	
	\subsection{Image restoration} 
	Image restoration involves the process of improving the quality of degraded images. Restoration methods can be grouped into denoising~\cite{buades2005review}, deblurring~\cite{richardson1972bayesian,lucy1974iterative}, super-resolution~\cite{park2003super}, compression artifact removal~\cite{shen1998review}, etc. depending on the kind of degradation, although they share many similarities. Lately, deep learning methods have been successful for image restoration tasks. Some of these methods can be applied to any degradation~\cite{chen2016trainable,zhang2018residual_restoration} while others are specific to the degradation (deblurring~\cite{nah2017deep,chen2019blind}, super-resolution~\cite{shi2016real,zhang2018residual}, denoising~\cite{zhang2017beyond,zhang2018ffdnet} and compression artifact removal~\cite{dong2015compression}). More recently, image restoration algorithms based on generative adversarial networks (GANs) have become popular owing to their improved performance (super-resolution~\cite{ledig2017photo, wang2018esrgan}, compression artifact removal~\cite{galteri2019deep, zhao2019compression} and deblurring~\cite{kupyn2018deblurgan}).
	
	A compressed image can be processed using a restoration method before using it for inference to improve its performance; although our analysis reveals that this is a sub-optimal approach. Galteri \textit{et al.}~\cite{galteri2019deep} propose a GAN-based restoration network to correct JPEG compression artifacts. They also evaluate different restoration algorithms on the basis of the performance of restored images on a trained object detection network. They show that their GAN-based algorithm performs better than other methods compared in the paper. Our analysis provides an explanation for this observation. 
	
	\subsection{Domain adaptation.} 
	
	Domain adaptation~\cite{wang2018deep} is a problem motivated by the lack of sufficient annotations. Typically, domain adaptation methods leverage the abundant annotated data available from a different yet related domain (called as the source domain) to improve performance on the domain of interest (target domain), where there is a lack of annotated data. Examples of source and target domains include synthetic images vs real images, images in the wild vs images on a webpage, etc. Domain adaptation methods can be divided into unsupervised~\cite{ganin2014unsupervised,bousmalis2017unsupervised}, semi-supervised~\cite{saito2019semi} and supervised~\cite{chen2012marginalized,glorot2011domain,tzeng2017adversarial,ganin2016domain,hoffman2017cycada} categories depending on the quantity of available data (and annotations) in the target domain. Approaches for domain adaptation can be categorized into latent feature alignment using autoencoders~\cite{chen2012marginalized,glorot2011domain}, adversarial latent feature alignment~\cite{tzeng2017adversarial,ganin2016domain,murez2018image,hoffman2017cycada} and pixel-level adversarial alignment~\cite{bousmalis2017unsupervised,hoffman2017cycada}.
	
	The scenario when only compressed images are available at training time, with original images available at test time is related to domain adaptation. Dataset restoration, our proposed method for this scenario, corrects the covariate shift and domain adaptation algorithms account for domain shift in some form. Probably, the closest domain adaptation method to dataset restoration is an unsupervised method that addresses alignment only at pixel-level~\cite{bousmalis2017unsupervised}. However, an important distinction is that domain adaptation tackles the problems arising due to lack of annotations for the images in the target domain, while for us the concern lies in the non-availability of images themselves. Thus, we study the effectiveness of dataset restoration using an external dataset for training and also by varying the number of original training images.
	
	\section{Learning and inference with compressed images}
	\label{sec:methods}
	
	\subsection{Problem definition}
	
	We are concerned with downstream understanding tasks where we want to infer from an input image $\mathbf{x}\sim p_X\left(\mathbf{x}\right)$, the corresponding semantic information $\mathbf{y}\sim p_{Y\vert X}\left(\mathbf{y}\right)$. In the rest of the paper, we will assume that $\mathbf{y}$ is a semantic segmentation map, but our approach can also be applied to other semantic inference tasks, such as image classification or object detection. The objective is to find a parametric mapping $\phi:\mathbf{x} \mapsto \mathbf{y}$ by supervised learning from a training dataset $X_{tr}=\left\{\left(\mathbf{x}^{(1)}_{tr},\mathbf{y}^{(1)}_{tr}\right),\ldots,\left(\mathbf{x}^{(N)}_{tr},\mathbf{y}^{(N)}_{tr}\right)\right\}$, where each image $\mathbf{x}^{(i)}$ has a corresponding ground truth annotation $\mathbf{y}^{(i)}$. The mapping is typically implemented as a deep neural network. The performance of the resulting model is evaluated on a test set $X_{ts}=\left\{\left(\mathbf{x}^{(1)}_{ts},\mathbf{y}^{(1)}_{ts}\right),\ldots,\left(\mathbf{x}^{(M)}_{ts},\mathbf{y}^{(M)}_{ts}\right)\right\}$. Under conventional machine learning assumptions,  $X_{tr}\sim p_X\left(\mathbf{x}\right)$ and $X_{ts}\sim p_X\left(\mathbf{x}\right)$, \textit{i.e.} both training and test sets are sampled from the same underlying distribution $p_X$.
	
	In our setting, we consider that $X_{tr}$ \textrm{and/or} $X_{ts}$ undergo a certain degradation $\psi:\mathbf{x} \mapsto \hat{\mathbf{x}}$. In our case, the degradation is related with the lossy compression process necessary to transmit the image to the remote location where the actual training or inference takes place; and so we have  $\hat{\mathbf{x}}=\psi\left(\mathbf{x}\right)=g\left(f\left(\mathbf{x}\right)\right)$, where $f\left(\mathbf{x}\right)$ is the image encoder, $g\left(\mathbf{z}\right)$ is the image decoder\footnote{We only consider lossy compression, since in lossless compression $\hat{\mathbf{x}}=\mathbf{x}$.} and $\mathbf{z}$ is the compressed bitstream. The result $\hat{\textbf{x}}$ is the reconstructed image, which follows a new distribution $p_{\hat{X}}$ of degraded images, \textit{i.e.} $\hat{\mathbf{x}}\sim p_{\hat{X}}\left(\mathbf{x}\right)$. Note that parallels can be drawn from the arguments in this section for other image degradations such as blur, downsampling, noise, color and illumination changes, etc. 
	
	Lossy compression is characterized by the distortion $D\left(\mathbf{x},\hat{\mathbf{x}}\right)$ of the reconstructed image and the rate $R\left(\mathbf{z}\right)$ of the compressed bitstream. The encoder and decoder are designed to operate around a particular rate-distortion (R-D) tradeoff $\lambda$, either by expert crafting in conventional image compression, or by directly optimizing parameters of a deep neural network. 
	
	\figuredegradationexamples
	
	\subsection{Covariate shift}
	The \textit{covariate shift} problem precisely occurs when the underlying distributions of training and test data differ, \textit{i.e.}  $X_{tr}\sim p_{X_{tr}}$ and $X_{ts}\sim p_{X_{ts}}$ with $p_{X_{tr}}\neq p_{X_{ts}}$. This leads to sub-optimal performance because the model is evaluated on a data distribution different from the one it was optimized for. While covariate shift is often found in machine learning (e.g. training with synthetic data and evaluating on real), in our case, this problem is a consequence of lossy compression and it increases severely as the rate decreases. The drop in performance is related to the degree of covariate shift, which could be seen as the divergence between distributions $d\left(p_{X_{tr}}, p_{X_{ts}}\right)$. In the conventional machine learning setting without compression, there is no covariate shift, since $X_{tr}\sim p_{X}$ and $X_{ts}\sim p_{X}$, nor when both training and test set are compressed with the same method and at the same rate, since $X_{tr}\sim p_{\hat{X}}$ and $X_{ts}\sim p_{\hat{X}}$. However, covariate shift exists in the other two configurations, namely CO and OC (see Table~\ref{tab:nomenclature}).
	
	The degradation due to lossy compression can be observed clearly in Fig.~\ref{fig:feature_vis}b,  when comparing the original captured image and the image after compression. This also gives an idea of the difference between the original domain and the domain induced by compression. It has images with lesser details which also suffer from blurring and coding artifacts. More examples are shown in Fig.~\ref{fig:degradationexamples} for the two compression methods (MSH and BPG), with the images compressed at a similar rate. It can be seen that degradations are consistent yet with some differences (e.g. blocky artifacts for BPG, more blurred in MSH).
	
	\subsection{Semantic information loss}
	Covariate shift explains how compression impacts the downstream task when data at training and test time are compressed unequally. 
	Another factor that impacts task performance arises from compression, resulting in \textit{semantic information loss}. By the semantic information present in an image we refer \textit{only} to what is relevant to the downstream task. Thus, by definition, semantic information loss is task dependent. Continuing with our example from the introduction (Fig.~\ref{fig:feature_vis}b), the letters in the license plate of the car plays little to no role in establishing the presence of a car in the image. Thus, the exact letters are not relevant semantic information for the task of car detection. However, if the task is license plate recognition, the letters are an integral part of semantic information. Compression causes semantic information loss as it makes the compressed image devoid of some semantic attributes present in the original image. The loss of letters on the plate in the compressed image (Fig.~\ref{fig:feature_vis}b) is evidence of semantic information loss (when the task is license plate detection). 
	
	Further evidence of semantic information loss can be found in Fig.~\ref{fig:degradationexamples}, since the degradation often removes details and textures, blends small objects together via blur and lack of contrast, and introduces confusing artifacts, preventing us from recognizing small objects at all (e.g. individual pedestrians), and making larger objects more difficult to recognize due to the loss of discriminative  details and textures (e.g. tree leaves). Only in retrospective, after observing the original undistorted crop, we can infer the small objects in the distorted image. Similarly, a semantic segmentation model will struggle to recognize them, or directly fail when the semantic information has disappeared completely (e.g. license plate number).
	
	Let $Y$ be a random variable that represents the semantic information in the original image, $X$. For instance, if the task is semantic segmentation, $Y$ would take values from the set of semantic maps of images. Mathematically, we formulate semantic information loss, $S$, in the compressed images, $\hat{X}$, using mutual information, $I$, as follows: $S_{Y}(X, \hat{X}) = I(X, Y) - I(\hat{X}, Y)$. Predictably, $S_{Y}(X, \hat{X})$ is non-negative, since $\hat{X}$ is produced from $X$ via the map $\psi$ and thus, we have $I(X, Y) \ge I (\hat{X}, Y)$ as a consequence of the data processing inequality. 
	
	\subsection{Training/test configurations, application scenarios and related work}
	\tabletraintestconfigurations
	
	Now, we focus on several training/test configurations and provide examples of real world scenarios (summarized in Table~\ref{tab:nomenclature}). A configuration is defined by the pair $\left(X_{tr},X_{ts}\right)$, with $X_i \sim p_X$ represented as O and $X_i \sim p_{\hat{X}}$ represented as C. Thus, the conventional machine learning setting (\textit{i.e.} without compression) corresponds to OO, and the configuration of Fig.~\ref{fig:feature_vis}a is CO, since $X_{tr}\sim p_{\hat{X}}$ and $X_{ts}\sim p_X$. The former does not suffer from semantic information loss nor covariate shift, while the latter does suffer from both. The CO configuration can also be generalized to other scenarios involving on-board analysis\footnote{Often called \textit{on-board perception}, but we prefer \textit{on-board analysis} to avoid confusion later.} where data capture and inference takes place in the device and the training in a server (e.g. autonomous cars, unmanned aerial vehicles and other robotic devices).
	
	\figuremcc

	The configuration OC involves training in the server with the original images, while inference is performed with compressed images, leading to semantic information loss and covariate shift. This is often the case when the capturing device has limited resources to perform complex analysis (e.g. smartphone), but can compress and send the content through a communication channel, and then receive back the results of analysis (e.g. predicted class, bounding box, segmentation map). Fig.~\ref{fig:mcc} illustrates the paradigmatic scenario of (mobile) cloud computing~\cite{hayes2008cloud,zhu2011multimedia,fernando2013mobile}. Another example of OC configuration is distributed automotive perception~\cite{lohdefink2019gan}, where the sensor module compresses the captured image and transmits it through the automotive bus system to the perception module where the downstream tasks are performed.
	
	The configuration CC appears in the previous scenario when training images are also compressed, and at the same rate as test images. In this case, both training and test images suffer from semantic information loss, but there is no covariate shift since both are sampled from the same $p_{\hat{X}}$.  
	
	\textbf{Compression before training.} We can remove the covariate shift from the configuration OC by transforming it into CC. This can be achieved by compressing the training data at the same rate and we refer to this adaptation approach as \textit{compression before training} (see Fig.~\ref{fig:adaptation_methods}a). Naturally, since CC is unaffected by covariate shift unlike OC, we expect the model with configuration CC to outperform configuration OC.
	
	As a downside of the process of compression before training, semantic information loss is additionally introduced into the training set. However, the presence or absence of semantic information loss in the test images is a major factor, while it is not the case with the training images. The segmentation network is trained using the entire set and if some class information is lost in a particular training image, its presence in other training images can compensate for it. Thus, we can usually get away with introducing some semantic information loss in the training set. In contrast, as the segmentation network is evaluated on individual images in the test set, the performance suffers critically by the presence of semantic information loss.
    
	\textbf{Compression before inference.} Similarly, we can also transform configuration CO into CC by compressing the test images. We refer to this process as \textit{compression before inference}  (see Fig.~\ref{fig:adaptation_methods}b). While this process allows us to correct the covariate shift due to compression, it also introduces semantic information loss at test time. The introduction of semantic information loss in the test is critical and can cause the performance of configuration CC to be even worse than the configuration CO \textit{at times} (as shown in Fig.~\ref{fig:feature_vis}c). Moreover, compression before inference requires installing a full compression encoder and decoder module on-board prior to the downstream task, resulting in a  significant computational penalty in the deployed system.
	
	\section{Dataset restoration}
	\label{sec:restoration}
	
	\subsection{Proposed approach}
	Motivated by the two limitations mentioned above, we propose \textit{dataset restoration} as an alternative approach that alleviates covariate shift without inducing semantic information loss in the test data (in contrast to compression before inference). The key idea is to adapt the training dataset using adversarial image restoration, and use the adapted dataset as actual training data for the downstream task (see Fig.~\ref{fig:adaptation_methods}c). In this way, the on-board analysis module can exploit all the information available in the captured image. Another important advantage is that adaptation takes place only in the server, and the resulting model can be readily and seamlessly deployed in the device with the same hardware, therefore without requiring to install any additional hardware nor increasing the inference cost.
	
	We now recall that a great deal of degradation is related to the loss of texture in the decoded image and the appearance of compression artifacts (these two factors are clearly apparent in Figs.~\ref{fig:feature_vis}b~and~\ref{fig:degradationexamples}). Thus, our goal is to find an appropriate image restoration technique that could learn from a given set of examples and provide us a way to remove the artifacts and recover texture in the images. 
	
	Our restoration module is based on adversarial image restoration, where a generative adversarial network (GAN)~\cite{goodfellow2014generative} conditioned on the degraded image is employed to improve the image quality. A GAN is based on two networks competing in an adversarial setting. The generator takes the input image and outputs the restored image. The discriminator observes real and restored images and it is optimized to classify between real and restored images. The generator, in contrast is optimized to fool the discriminator, and indirectly improves the quality of the restored images. Through the process, the generator learns to remove compression artifacts and replace unrealistic textures by realistic ones that could be used by the discriminator to identify the restored images. The architecture of GAN is based on the one proposed in \cite{wang2018high} (for image-to-image translation~\cite{isola2017image}), which has a generator and multiple discriminators (see Appendix~\ref{appendix:air} for details). 
	
	During the process of dataset restoration, we use our trained generator to restore individually every image in the training dataset for the downstream task. Examples of some of the restored images can be found in Figs.~\ref{fig:feature_vis}b~and~\ref{fig:degradationexamples}. While not being able to restore lost semantic information (e.g. the same individual pedestrians), the restored images look sharper, have fewer artifacts and blurred regions are enhanced with hallucinated textures that resemble the real images. As such, the shift with respect to the distribution of original images, on which the trained model is to be evaluated, is reduced. Table~\ref{tab:nomenclature} includes two new configurations OR and RO, where R refers to restored images. 
	
	\figureaadaptationmethods
	
	\subsection{Adversarial restoration, covariate shift and perceptual index}
    
    Perceptual image quality is often assessed using subjective evaluations where human subjects are presented with pairs of images where one of them is degraded (generated through some artificial processing, such as compression or restoration) and the other is a real, not-degraded image. The perceptual quality is (inversely) proportional to the probability of correctly selecting the real image. Blau and Michaeli~\cite{blau2018perception} show that this probability, and therefore the perceptual quality, can be related to the divergence $d(p_X, p_Z)$ (in principle it could be \textit{any} probabilistic divergence) between the distribution of real images $p_X$ and the distribution of generated images $p_Z$. This probabilistic divergence is termed as \textit{perceptual index}. The lower the value of the perceptual index, the higher is the quality of the image. 
    
    In practice, collecting human opinions is expensive and often infeasible. Hence, \cite{blau2018perception} proposes other practical methods to estimate the perceptual index. Specifically, since the task of a discriminator neural network is precisely to distinguish between real and artificial images, its success rate could be used as perceptual index. With this point of view, training a GAN can be seen as decreasing the perceptual index of generated images by decreasing the perceptual index measured by the discriminator. However, a discriminator needs to be trained for every experiment, and also requires many images. As a more practical solution, \cite{blau2018perception} also suggests that Blind Image Quality Assessment methods can be a suitable proxy for the perceptual index, since these methods are trained to predict the actual human opinion scores in image quality assessment tests.
	
	With the interpretation of perceptual index as a divergence with respect the distribution of real images, we observe an interesting relation between perceptual index and covariate shift, which explains why adversarial image restoration is an appropriate approach (compared to non-adversarial). In the CO configuration, the training images $X_{tr}$ are compressed and therefore follow $p_{\hat{X}}$, while the test images $X_{ts}$ follow $p_X$. Thus, the covariate shift can be quantified as $d(p_X, p_{\hat{X}})$, where $d$ denotes a probabilistic divergence. Note that this quantity is essentially the perceptual index of compressed images. Therefore, in the case of CO configuration, the covariate shift in the configuration corresponds to the perceptual index of the training set. Now, an important conclusion from~\cite{blau2018perception} is that perception and distortion are at odds with each other, and that there exists a limit beyond which perception and distortion cannot be reduced simultaneously (see Fig.~\ref{fig:illus_percep_dist}). Thus, the effect of dataset restoration (\textit{i.e.} moving from CO to RO) is to lower the covariate shift at the expense of increasing distortion, provided we are close to the perception-distortion limit.
	
	\figureilluspercepdist
	
	We are ultimately interested in the implications on the performance of the downstream task, semantic segmentation in particular. Our analysis in the previous section reveals that the task performance is greatly dependent on covariate shift and semantic information loss. Reducing the perceptual index is therefore more crucial than training with images of low distortion, as a lower perceptual index corresponds to a lower covariate shift. Hence, we use adversarial restoration and decrease the perceptual index at the cost of increased distortion.

	Are all image restoration approaches helpful? We argue that only adversarial image restoration are suitable, since they explicitly minimize the perceptual index through the discriminator and consequently the covariate shift with respect to the captured images. In contrast, non-adversarial image restoration methods do not necessarily reduce the perceptual index. Typically, these methods try to further decrease the distortion and this can be counter-productive as perception and distortion are at odds near the limit.

	\subsection{Training data for dataset restoration}
	Training the GAN for dataset restoration requires original images as the discriminator is tasked to distinguish between the image output from the generator and original images. We consider two cases depending on the data available:
	
	\begin{enumerate}[\indent a.]
	    \item \textbf{Privileged dataset.} We assume the availability of some amount of original images from the \textit{same distribution} (\textit{i.e.} $p_X$) that can be used to train the restoration network (e.g. collected with lossless compression). These privileged images are generally much more expensive to collect than the usual (lossily) compressed images. Note that for configurations RO and OR, the restoration network is trained using privileged data.
	    \item \textbf{Auxiliary dataset.} We use an external dataset $Z$ with uncompressed images, preferably from a similar domain. This option has typically zero cost, since we can leverage publicly available image restoration datasets, or even directly use a publicly available adversarial restoration model. We denote as AO the configuration RO when the restoration network is trained using an auxiliary dataset.
	\end{enumerate}
	
	The images in $Z$ follow a distribution $p_Z\neq p_X$. Thus, training the restoration network with the auxiliary dataset  has the drawback of suffering from certain \textit{domain shift}, which does not occur in the privileged dataset. However, the degradations and artifacts that a restoration network restores tend to be local and low-level, which are largely shared across different domains. In general, dataset restoration with an auxiliary dataset is already effective and a budget option, while a privileged dataset without domain shift is more effective, but incurs the additional cost of collecting it.
	
	\section{Experiments}
	\label{sec:results}
	
	\subsection{Experimental settings}
	\textbf{Datasets.} We evaluate our methods on three datasets:
	
	\textit{Cityscapes}~\cite{cordts2016cityscapes} is a popular dataset in autonomous driving, and contains 5000 street images (2975/500/1525 for training/validation/test sets) of which training and validation have pixel-level segmentation maps annotated with 19 different concepts, including objects and ``stuff''. We use the annotated sets to train (training set) and evaluate semantic segmentation (validation set). It also contains another 20000 images with coarse annotation. We ignore these annotations and use a subset of 2000 images to train the deep image compression model (\textit{i.e.} MSH~\cite{balle2018variational}) and the image restoration methods.
	
	\textit{INRIA Aerial Images Dataset}~\cite{maggiori2017dataset} contains aerial images of diverse urban settlements with segmentation maps with two classes (building and background). The dataset consists of aerial images from 10 cities with 36 images per city. Annotations are provided for 5 of these cities and the segmentation models were trained on 4 cities and evaluated on 1 from these. The images from the other 5 cities were used for compression and restoration. 
	
	\textit{Semantic Drone Dataset}~\cite{mostegel2019semanticdronedataset} contains 400 high resolution images captured with an autonomous drone at an altitude of 5 to 30 meters above ground, and their corresponding annotated segmentation maps (20 classes). The 400 publicly released images were resized from a resolution of 6000x4000 to 3000x2000. The segmentation models were trained on 265 images and evaluated on 70 images while the remaining 65 images were used for the compression and restoration models. Each image was further split into 12 patches each with dimension of a 1200x800. All metrics are calculated on these patched images.
	
	\textbf{Compression methods.} We use two state-of-the-art image compression methods. The Better Portable Graphics (BPG) format~\cite{bellard2017bpg} is based on a subset of the video compression standard HEVC/H.265~\cite{sullivan2012overview} and is the state-of-the-art in non-deep image compression. The Mean Scale Hyperprior (MSH)~\cite{balle2018variational,minnen2018joint} is a state-of-the-art deep image compression method, based on an autoencoder whose parameters are learned to jointly minimize rate and distortion at a particular tradeoff $\lambda$, \textit{i.e.} $\min R+\lambda D$. MSH models were pretrained for 600k iterations on the CLIC Professional Dataset\footnote{https://www.compression.cc/2019/challenge/} with \textit{MSE} loss. Appendix~\ref{appendix:msh} contains details of the model architecture. 
	
	\textbf{Restoration methods.} Our adversarial restoration architecture for the proposed dataset restoration method is based on FineNet~\cite{akbari2019dsslic}. FineNet is an adaptation of Pix2PixHD~\cite{wang2018high}, a popular GAN architecture used for a broad range of image-to-image translation problems. Refer to Appendix~\ref{appendix:air} for more details. Further, when comparing adversarial and non-adversarial approaches, we use Residual Dense Network~\cite{zhang2018residual_restoration} (Appendix~\ref{appendix:rdn}) as a representative method of non-adversarial restoration.
	
	\textbf{Segmentation.} For the downstream task we use the state-of-the-art semantic segmentation method DeepLabv3+~\cite{chen2018encoder}. The model is trained using the same procedure mentioned in the paper. We use an output stride of 16 and perform single scale evaluation.
	
	\textbf{Metrics.} The quality of the inferred semantic segmentation map is evaluated using the mean intersection over union (mIoU, the higher the better). For image compression we measure rate in bits per pixel (bpp, the lower the better) and the distortion in PSNR (in dB, the higher the better). 
	
	\subsection{Cityscapes}
	\figurecityscapesrdcurvesnew
	
	\textbf{Rate-distortion curves.} We first characterize the rate-distortion performance for the two compression methods in our experiments, and the impact of the proposed restoration approach on them (see  Fig.~\ref{fig:cityscapes_rd_curves}). The curves sweep the whole range, from low to high quality images. As expected, the distortion decreases (PSNR increases) with rate. We observe that MSH performs significantly better than BPG on Cityscapes, \textit{i.e.} it produces images with lower average distortion at similar rate. Due to the perception-distortion tradeoff, restoration leads to increase in the average distortion, which can be observed in Fig. ~\ref{fig:cityscapes_rd_curves}. Interestingly, once the images are restored, images compressed with MSH have marginally higher distortion than those compressed with BPG. 
	
	\figurecityscapessegmentationperformance
	
	\tablemiouperclassfinal
	
	\textbf{Segmentation performance.} We evaluate the segmentation performance under seven different configurations (\textit{i.e.} OO, CO, RO, AO, CC, OC and OR). The results are shown in Fig.~\ref{fig:cityscapes-segmentation-performance}. 
	
	For the Cityscapes dataset, we observe that the model with configuration CO outperforms the model with configuration CC, which shows that correcting covariate shift by compression before inference can potentially result in lowering the performance. Table~\ref{tab:cityscapes-per_class_miou} shows the performance per class. We see that the mIoU of classes representing small objects\footnote{We consider classes \textit{person}, \textit{rider}, \textit{motorcycle}, \textit{bicycle}, \textit{pole}, \textit{traffic light}, \textit{traffic sign} as small objects and the remaining classes as big objects.} is significantly lower in the configuration CC when compared to CO. Since smaller objects are relatively easy to lose by compression, the observation confirms that the introduction of semantic information loss in the test set from the process of compression before inference is responsible for the decrease in performance.
	
	The proposed dataset restoration approach, RO is able to improve  1.4-4.8\% on the configuration CO, and we achieve close to optimal performance (77\% mIoU) requiring only 0.13 bits per pixel. Thus, lossy compression can result in huge storage savings during data collection when compared to lossless image compression. For the same budget required to collect the 2975 training images with MSH at 0.13 bpp, using lossless compression (PNG in our experiments, resulting in 9 bpp) we would have collected only 42 images, which is clearly insufficient to train the segmentation network. Furthermore, for the same performance, dataset restoration effectively reduces the required budget (for example, configuration CO achieves 75.64\% with 0.128 bpp, while RO approximately requires around 0.07 bpp).

    So far in this section, we have considered the availability of privileged data. This allowed us to avoid any domain shift effects that arise when restoration network is trained with auxiliary data. In the following subsections, however, we consider different training sets for restoration.
 	
	\textbf{Restoration network (auxiliary data).}
    As described in Section~\ref{sec:restoration}, we train the restoration network with an auxiliary dataset to evaluate the effectiveness of dataset restoration when privileged data is not available. We use the images from the front center camera of the Ford Multi-AV Seasonal Dataset~\cite{agarwal2020ford} as an auxiliary dataset. 
	
	Note that, while both being driving datasets, there are many differences between the Ford dataset and the Cityscapes dataset. Cityscapes is a rich and diverse dataset collected from multiple cities in Germany. The images in the Ford dataset are obtained by driving a car along a single route in Michigan, USA and hence it lacks diversity. The camera sensors and the resolution of the images differ as well.
	
    \figurerestorationdatarequirednew
 	
    The configurations AO, RO, CO and CC are compared in Fig.~\ref{fig:cityscapes-segmentation-performance} and  Table~\ref{tab:cityscapes-per_class_miou}. Despite all the aforementioned differences between the datasets, we observe only a small decrease in terms of performance when AO is compared to RO. The configuration AO still performs significantly better than the baselines, CO and CC. Note that the results for AO depend on the auxiliary dataset collected which could be improved with a better auxiliary dataset.
    
	\textbf{Amount of privileged data.} Since collecting privileged data is expensive, the amount of collected images is an important factor. This privileged data is readily available in the server side, and could be leveraged to train the segmentator, restoration network or both. In the following experiments, we consider three different amounts of privileged data, 12.5\% (373 images), 25\% (745 images) and 50\% (1489 images) of the size of the segmentation dataset (2975 images).
	
	First, we evaluate the segmentation network trained solely with different amounts of original images from the privileged dataset. These configurations are oO - 12.5\%, oO - 25\% and oO - 50\%.

    Next, we train the restoration network with the three different amounts of privileged data mentioned above. The segmentation network is then trained on the images obtained after restoring the compressed data using the respective restoration models. These configurations are rO - 12.5\% (373 images), rO - 25\% (745 images) and RO (1489 images)\footnote{The configuration RO is the same as rO - 50\%.}.
    
    Fig.~\ref{fig:restoration_data-required} shows the results for the various configurations mentioned above. For all the different amounts of privileged data considered, dataset restoration is able to improve the performance over CO. When the privileged data collected is 12.5\% of the original images, the original images themselves are insufficient to provide a good performance as the configuration rO - 12.5\% performs better than oO - 12.5\% at all rates. When 25\% privileged data is available, the picture is a bit different. At low bitrates, training the segmentation network with privileged data is sufficient, while at higher bitrates restoration is beneficial. At rates greater than 0.087 bpp, rO - 25\% performs better than oO - 25\%. However, when privileged data is available in copious amounts (e.g. 50\%), restoration performs worse even at high bitrates and privileged data can be directly used for training the segmentation network. Thus, to understand the benefits of restoration better, we consider training the segmentation network with both restored and original images, and then compare the data collection cost against segmentation performance for all the  configurations in the following subsections.

	\textbf{Hybrid training sets.} Since privileged and compressed images are available in the server, we now consider hybrid training sets for the segmentation network where 12.5\% of the images (373 images) are privileged and the remaining are compressed or restored, and evaluate on original images (configuration o+c O or o+r O, respectively). Since we intend to evaluate the segmentation network on original images, we emphasize the contribution of the original images in the loss during training in order to achieve a higher performance. For all the experiments reported in Table~\ref{tab:hybridtrainingsets}, the loss from original images and compressed (or restored) images are weighed in the ratio of 5:1 (empirically determined). 
    
    \tablehybridtrainingsets

    We observe that the models trained with hybrid training sets perform better than the individual components of the mixture; \textit{i.e.}, configuration o+c O performs better than CO and oO, while o+r O performs better than oO and rO. Between the models of the two mixtures, we see that dataset restoration can still help performance in the case of BPG. For MSH, dataset restoration may not be necessary. 
    
    \textbf{Segmentation performance vs. data collection cost.} This work is motivated by the need to reduce the data collection cost. However, privileged, compressed and auxiliary images have different costs associated with their collection (high, medium-low, and zero, respectively). In order to provide a complete picture, in Fig.~\ref{fig:costperformance}, we plot the performance of all the configurations considered thus far against the total cost involved in collecting the required data (for training restoration and segmentation networks). The cost is reported in terms of the percentage of the total cost for the OO configuration. We see that in the low cost region ($\le$ 1\%), dataset restoration with auxiliary data provides the best performance, since auxiliary data involves no cost. When a higher performance is needed, privileged data needs to be collected and thus a higher cost is incurred. In such cases, the cost due to the rate of compression is dominated by the cost in collecting the privileged data. Hybrid training sets, particularly o+c O with MSH and o+r O with BPG, result in the best performance in the high cost region. When the budget is very high (e.g. 50\% or higher), lossless compression can be directly used to collect images, although our motivation in the first place is precisely to avoid those very high budget requirements.
    
    \figurecostperformance
	
	\tableperceptualindices
	\figureadversarialvsnonadversarial
    \tablenewrdnresults

	\textbf{Adversarial vs. non-adversarial restoration.} In order to show the connection between perceptual index and adversarial image restoration, we compute the perceptual index of the original, compressed and restored images using the blind quality assessment method HOSA~\cite{xu2016blind} (as described in~\cite{blau2018perception}). We use Residual Dense Networks~\cite{zhang2018residual_restoration} as a representative method of non-adversarial image restoration. RDN is trained with the objective of increasing the PSNR (RDN - PSNR) or MS-SSIM~\cite{wang2003multiscale} (RDN - MS-SSIM). Refer to Appendix~\ref{appendix:rdn} for more details. Note that adversarial restoration is able to achieve a perceptual index close to that of the original images while RDN does not affect the perceptual index significantly.
	
	\figureimagesfeatcorr
	
	While the previous experiment shows that adversarial restoration does improve the perceptual quality of images, we are ultimately interested in the segmentation task. We use the model trained with the original images and extract features from a shallow layer. Features are obtained from original images, compressed images and restored images for comparison. As compression affects the low-level information in images by degrading the texture, adding blur and other alien artifacts, a shallow layer is selected to observe how these low-level differences between the images are reflected in semantic features. Since we use the same model to extract the different features, they are all aligned in the channel dimension. We want to measure the level of alignment in activation, so we plot the average activation (over the validation set of  Cityscapes) value for each channel of original, compressed or restored versus the original one (see Fig.~\ref{fig:images_feat_corr}). Obviously, the points corresponding to original images lie on the identity line, while many channels from compressed images are clearly not aligned. Adversarial restoration manages to bring back the features to the identity line, while non-adversarial restoration has little effect. This shows that adversarially restored images are not only perceptually closer to the real images, but also semantically more correlated.
	
	Further, Table~\ref{tab:rdn_results} compares the performance of different restoration methods in the configuration RO. Along with segmentation performance in terms of mIoU, distortion measures of PSNR and MS-SSIM\footnote{MS-SSIM (dB) is calculated from the standard MS-SSIM value (range [0, 1]) as follows: MS-SSIM (dB) $= \; -10 \log_{10}(1 -$ MS-SSIM$)$. MS-SSIM is presented in terms of this logarithmic scale for better distinction.} are also reported. Adversarial restoration results in a far better segmentation performance when compared to RDN. 
	
	\textbf{Efficiency.} Table~\ref{tab:efficiency} reports the inference times for different configurations. The time for encode-decode times for MSH and the segmentation time for DeepLabv3+ were measured on a Quadro RTX6000 GPU, while the encode-decode times for BPG was measured on a Intel Xeon(R) E5-1620 v4 CPU\footnote{We used the software implementation of BPG from https://github.com/mirrorer/libbpg using the default options.}. Inference using dataset restoration (configuration RO) is faster than compression before inference (configuration CC) by 26\% and 66\%, when the compression method used is MSH and BPG respectively.
	
	\tableefficiency
	
	\subsection{INRIA Aerial Images Dataset}
	
    \figureaidsegmentationperformance

	\textbf{Segmentation performance.} Fig.~\ref{fig:aid-segmentation-performance} depicts the segmentation results obtained on the INRIA Aerial Images Dataset (AID) for the same six configurations mentioned in Section B. We observe that the model with configuration CO performs better than CC when the compression method used is BPG. However, the same cannot be said for MSH. We hypothesize that the use of MSH, which causes smoothing artifacts, destroys the discriminative features and we are unable to learn a segmentation model capable of taking advantage of these features in the original image. This is especially critical with the AID since there are only a few features that discriminate a building from the background. 

	The proposed approach of RO performs consistently better than both these configurations with gains up to 3.9 \% mIoU. Fig.~\ref{fig:aidexample} shows a portion of an image from the dataset along with the segmentation maps predicted by various models.
    
    Interestingly, contrary to the results on the Cityscapes dataset, the performance of configuration OR does not improve over OC. This suggests that the process of restoration causes further damage (over compression) to the discriminative features used by the model trained on the original images for its prediction. 

	\figureaidexample
    
	\subsection{Semantic Drone Dataset}

    \figuresddsegmentationperformance	

	\textbf{Segmentation performance.} The segmentation results obtained on the Semantic Drones Dataset (SDD) are shown in Fig.~\ref{fig:sdd-segmentation-performance}. We observe that the models with configuration CC outperform CO consistently for both MSH and BPG. We attribute this result to the lack of semantic information loss in this dataset. A typical object from the SDD covers a significant portion of the image and compression does not significantly affect its recognizability. When we assume little to no semantic information loss, the effects of covariate shift are dominant and as such, the configuration CC performs better. This result shows that the performance of various configurations are dependent on the properties of the dataset. 
	
	The proposed approach of RO performs similarly to configuration CC and lies within $\pm 1.5\% $ mIoU of the configuration CC. The lack of significant semantic information loss in SDD affects the effectiveness of dataset restoration.
	
	\section{Conclusions}
	The rapid development in sensor quality and increasing data collection rate makes lossy compression necessary to reduce transmission and storage costs. By means of dataset restoration, we enable the incorporation of lossy compression for on-board analysis, greatly mitigating the drop in performance. Moreover, dataset restoration is a principled approach, based on our analysis of the various scenarios involving learning and inference with compressed images. This analysis framework involving covariate shift and semantic information loss can be further extended to other degradations like blur, noise, color and illumination changes, etc. 
	
	\appendices
	
	\section{Mean scale hyperprior}
	\label{appendix:msh}
	\figuremsharch
	Fig. \ref{fig:msh_arch} describes the architecture of the MSH~\cite{minnen2018joint} used in this paper.
	
	\section{Adversarial image restoration}
	\label{appendix:air}
	We use the FineNet from Akbari \textit{et. al}~\cite{akbari2019dsslic} (adapted from~\cite{wang2018high}) with slight changes for our image restoration module.
	
	Following the same notation in~\cite{akbari2019dsslic}, the generator architecture is written as
	$ c_{64}, \; d_{128}, \; d_{256}, \; d_{512}, \; 9 \times r_{512}, \; u_{256}, \; u_{128}, \; u_{64}, \; o_{3} $ where
	\begin{itemize}
		\item $c_k$: Conv: 7 × 7 x k, Instance Normalization, ReLU
		\item $d_k$: Conv: 3 x 3 x k / $\downarrow$ 2, Instance Normalization, ReLU
		\item $r_k$: Conv: 3 x 3 x k, Reflection padding, Instance Normalization, ReLU
		\item $u_k$: Conv: 3 x 3 x k / $\uparrow$ 2, Instance Normalization, ReLU
		\item $o_3$: Conv: 7 x 7 x 3, Instance Normalization, Tanh
	\end{itemize}
	
	We use two discriminators, as in ~\cite{akbari2019dsslic}, operating at two different scales. Akbari \textit{et al.} rescale the image to half the resolution while we do not. The discriminators act on the original resolution, $H \times W $ and $H/4 \times W/4$ resolution. Again following notation in ~\cite{akbari2019dsslic}, the discriminators have the following architecture,
	$ C_{64}, \; C_{128}, \; C_{256}, \; C_{512}, \; O_{1} $, where 
	
	\begin{itemize}
		\item $C_k$: Conv: 4 x 4 x k / $\downarrow$ 2, Instance Normalization, LeakyReLU
		\item $O_1$: Conv: 1 x 1 x 1
	\end{itemize}
	
	Let the captured image be $x$. The restored image, $ \bar{x} $ is obtained by adding the residual computed by the generator to the compressed image, $$ \bar{x} = \hat{x} + G(\hat{x}).$$
	
	All images are scaled to $[-1, 1]$.
	
	The loss function used for training are as follows:
	\begin{itemize}
		\item Generator, $G$: $$ L_{GAN}^{(G)} + 10 \cdot (2 \cdot  L_{1} + L_{VGG} + L_{MS-SSIM} + L_{DIST}) $$
		\item Discriminator, $D_i$: $ L_{GAN}^{(D_i)} $ 
	\end{itemize}
	
	$L_{GAN}^{(G)}$ is the sum of the standard GAN loss from each of the discriminator, \textit{i.e.}
	$$ L_{GAN}^{(G)} = \sum_{i=1}^{2}{- \log(D_{i}(\hat{x}, \bar{x}))}.$$
	
	\newcommand\norm[1]{\left\lVert#1\right\rVert_1}
	
	$$ L_{1} = \norm{\bar{x} - x}.$$
	$$ L_{MS-SSIM} = \textrm{\textit{MS-SSIM}}(\bar{x}, x).$$
	
	Let VGG denote a VGG-Net trained on the ImageNet dataset and $M_{j}$ denote the size of the output of the $j^{th}$ layer of VGG. The output of each of the 5 convolution blocks are considered for the VGG feature distillation loss, which is given by
	
	$$ L_{VGG} = \sum_{j=1}^{5} \frac{1}{M_{j}} \norm{VGG^{(j)}(\bar{x}) - VGG^{(j)}(x)}.$$
	
	Similarly, the features of the discriminators are also distilled for stable GAN training.
	
	$$ L_{DIST} = \sum_{i=1}^{2} \sum_{j=1}^{4} \frac{1}{N_{j}^{(i)}} \norm{D_{i}^{(j)}(\hat{x}, \bar{x}) - D_{i}^{(j)}(\hat{x}, x)}.$$
	
	The discriminators are trained using the standard GAN loss.
	
	$$L_{GAN}^{(D_i)} = \log(1 - D_{i}(\hat{x}, \bar{x})) + \log(D_{i}(\hat{x}, x)).$$
	
	We use a batchsize of 1 and train the GAN for around 135k iterations. Adam optimizer with $\beta_1 = 0.1$ and $\beta_2=0.9$ is employed. Initially, the learning rate is set to 0.0002 and is reduced by a factor of 10 after 80k iterations. 
	
	\section{Residual dense network}
	\label{appendix:rdn}
	We use the RDN architecture from \cite{zhang2018residual_restoration}. We ask the reader to refer to the paper for the architecture. The following hyperparameters are used: Global layers = 16, Local layers = 6, Growth rate = 32.
	
	We train the CAR model with the objective of maximising MS-SSIM or PSNR. The models are trained using 256x256 patches. A mini-batchsize of 1 is used and the model is trained for around 200k iterations. Adam optimizer is used with initial learning rate of 0.001 which is reduced by a factor 10 at 80k and 150k iterations.
	
	\ifCLASSOPTIONcompsoc
	\section*{Acknowledgments}
	\else
	\section*{Acknowledgment}
	\fi
	
	The authors thank Audi Electronics Venture GmbH, the Generalitat de Catalunya CERCA Program and its ACCIO agency for supporting this work. Luis acknowledges the support of the Spanish project RTI2018-102285-A-I00 and the Ramón y Cajal fellowship RYC2019-027020-I. Joost acknowledges the support of the Spanish project PID2019-104174GB-I00. Antonio acknowledges the support of project TIN2017-88709-R (MINECO/AEI/FEDER, UE) and the ICREA Academia programme.
	
	\ifCLASSOPTIONcaptionsoff
	\newpage
	\fi
	
	\bibliographystyle{IEEEtran}
	\bibliography{refs}
	
\end{document}